\documentclass{article}

\usepackage[preprint,nonatbib]{neurips_2025}

\usepackage[utf8]{inputenc} 
\usepackage[T1]{fontenc}    
\usepackage{url}            
\usepackage{booktabs}       
\usepackage{amsfonts}       
\usepackage{nicefrac}       
\usepackage{microtype}      
\usepackage{xcolor}         

\usepackage{multirow}
\usepackage{comment}
\usepackage{pifont}
\usepackage{graphicx}
\usepackage{amsmath}
\usepackage{amssymb}
\usepackage{wrapfig} 
\usepackage{tikz}
\usepackage{siunitx}
\usepackage{adjustbox}
\usepackage{enumitem}
\usepackage{wrapfig}

\usepackage{placeins}

\usepackage{colortbl} 
\definecolor{top1}{HTML}{FFA500}  
\definecolor{top2}{HTML}{FFCC70}  
\definecolor{top3}{HTML}{FFECB3}  
\definecolor{iccvblue}{rgb}{0.21,0.49,0.74}
\usepackage[pagebackref,breaklinks,colorlinks,allcolors=iccvblue]{hyperref}
\usepackage[capitalize]{cleveref}
\crefname{section}{Sec.}{Secs.}
\crefname{table}{Tab.}{Tabs.}
\crefname{figure}{Fig.}{Figs.}

\title{OpenMaterial: A Large-scale Dataset of Complex Materials for 3D Reconstruction}

\author{
  \textsuperscript{1}Zheng Dang\;\; \textsuperscript{2}Jialu Huang\;\; \textsuperscript{2}Fei Wang\;\; \textsuperscript{1}Mathieu Salzmann \\
  \textsuperscript{1}CVLab, EPFL, Switzerland\;\;\textsuperscript{2}Xi'an Jiaotong University, China \\
}

\begin{document}
\maketitle

\newcommand{\MS}[1]{\textcolor{red}{{\bf #1}}}
\newcommand{\ZD}[1]{{\color{blue}{\bf ZD: #1}}}
\newcommand{\zd}[1]{{\color{blue}{#1}}}
\newcommand{\HJL}[1]{{\color{green}{\bf HJL: #1}}}
\newcommand{\hjl}[1]{{\color{green}{#1}}}

\newcommand{\cmark}{\ding{51}}
\newcommand{\xmark}{\ding{55}}
\newcommand{\yes}{\textcolor{green}{\cmark}}
\newcommand{\no}{\textcolor{red}{\xmark}}
\newcommand{\synthetic}{\textcolor{red}{synthetic}}
\newcommand{\studio}{\textcolor{red}{studio}}
\newcommand{\wild}{\textcolor{green}{in-the-wild}}

\newcommand{\bD}{\mathcal{D}}
\newcommand{\bF}{\mathcal{F}}
\newcommand{\bG}{\mathcal{G}}

\newcommand{\imghe}{.18\textwidth}
\vspace{-2em}
\begin{abstract}
Recent advances in deep learning, such as neural radiance fields and implicit neural representations, have significantly advanced 3D reconstruction. However, accurately reconstructing objects with complex optical properties—such as metals, glass, and plastics—remains challenging due to the breakdown of multi-view color consistency in the presence of specular reflections, refractions, and transparency. This limitation is further exacerbated by the lack of benchmark datasets that explicitly model material-dependent light transport. To address this, we introduce OpenMaterial, a large-scale semi-synthetic dataset for benchmarking material-aware 3D reconstruction. It comprises 1,001 objects spanning 295 distinct materials, including conductors, dielectrics, plastics, and their roughened variants, captured under 714 diverse lighting conditions. By integrating lab-measured Index of Refraction (IOR) spectra, OpenMaterial enables the generation of high-fidelity multi-view images that accurately simulate complex light-matter interactions. It provides multi-view images, 3D shape models, camera poses, depth maps, and object masks, establishing the first extensive benchmark for evaluating 3D reconstruction on challenging materials. We evaluate 11 state-of-the-art (SOTA) methods for 3D reconstruction and novel view synthesis, conducting ablation studies to assess the impact of material type, shape complexity, and illumination on reconstruction performance. Our results indicate that OpenMaterial provides a strong and fair basis for developing more robust, physically-informed 3D reconstruction techniques to better handle real-world optical complexities. The dataset is available at~\href{https://huggingface.co/datasets/EPFL-CVLab/OpenMaterial}{link}.

\end{abstract}
\vspace{-2em}

\section{Introduction}
\vspace{-.5em}

Recovering the 3D geometry of a scene from images has been a long-standing and fundamental problem in computer vision. Successfully achieving such 3D reconstruction would facilitate numerous downstream tasks. For example, it can be leveraged in embodied AI and robotics, enabling efficient real-to-simulation workflows and supporting the creation of realistic, scalable simulation environments. It also plays a key role in building digital twins of the real world using 
mobile devices.

In recent years, the field of 3D reconstruction has seen significant progress, particularly via the development of neural radiance fields (NeRFs), which, although often targeting novel view synthesis, aim to internally extract a 3D scene representation. 
In this context, while the pioneering NeRF approach represents 3D scenes using a volumetric radiance field without explicitly modeling surfaces, subsequent approaches have incorporated implicit neural representations, such as deep signed distance functions (SDFs), to improve the accuracy of 3D surface reconstruction.
In particular, methods like Neuralangelo~\cite{Li23}, BakedSDF~\cite{Yariv23}, and Instant-NGP~\cite{Muller22} perform well on objects with diffuse surfaces. However, their performance diminishes when reconstructing surfaces of objects made from complex materials such as metals and glass. The reflective and refractive properties of these materials disrupt multi-view color consistency, a fundamental assumption in general-purpose 3D reconstruction algorithms, leading to reconstruction artifacts.

To address this challenge, specialized methods such as NeRO~\cite{Liu23b} have been designed to handle specular reflective surfaces using BRDF-based (Bidirectional Reflectance Distribution Function) physical loss functions.
While such methods demonstrate the benefits of integrating physical models of reflection into reconstruction algorithms, there remains ample room for improvement in reconstructing materials with all kinds of complex optical properties. In reality, an important limitation hindering progress in this area is the scarcity of high-quality datasets containing objects with diverse non-diffuse materials. Most existing datasets primarily contain diffuse materials, with non-diffuse materials constituting only a minor fraction of the overall material composition. The lack of extensive real-world datasets stems from the challenges and costs associated with acquiring and annotating high-fidelity 3D models of objects with complex optical properties.

In this paper, we bridge this gap by introducing OpenMaterial, a semi-synthetic dataset designed to facilitate material-aware 3D reconstruction. OpenMaterial features accurate simulations of material-light interactions, incorporating a diverse range of material types, geometric shapes, and lighting conditions. Unlike traditional datasets that approximate optical properties using a single-valued IOR, our dataset integrates 294 IOR spectra of non-diffuse materials, meticulously sourced from laboratory measurements. This enables accurate modeling of wavelength-dependent light-material interactions, resulting in significantly improved realism.

The dataset includes 1001 unique 3D geometries spanning a broad range of shapes, 295 materials categorized into conductors, dielectrics, plastics, and their roughened variants, and 714 High Dynamic Range Imaging (HDRI) maps representing diverse lighting conditions, ensuring a comprehensive evaluation framework for 3D reconstruction algorithms. Meanwhile, OpenMaterial provides multi-view images, 3D shape models, camera poses, depth maps, and object masks, making it the first large-scale dataset tailored for benchmarking material-aware 3D reconstruction. To validate the utility of OpenMaterial, we benchmark 11 SOTA 3D reconstruction and novel view synthesis methods. Furthermore, we conduct ablation studies with controlled variables to analyze the impact of material properties, geometric shapes, and lighting conditions on reconstruction performance.

Our contributions can be summarized as follows:
\vspace{-0.5em}
\begin{itemize}[left=0pt]
    \item We introduce OpenMaterial, a large-scale semi-synthetic dataset integrating laboratory-measured IOR spectra of materials and physics-based modeling to simulate light-material interactions. It encompasses diverse materials, shapes, and lighting conditions, with detailed annotations for in-depth study and algorithm evaluation.
    \item We benchmark SOTA algorithms for 3D reconstruction and novel view synthesis on our dataset, introducing a novel dimension of evaluation based on material types. We also conduct in-depth ablation studies to explore how material, shape, and illumination affect reconstruction accuracy.
    \item Upon acceptance, we will release all associated data and code, including dataset generation, benchmarking tools, and curated IOR spectra data of materials, to promote reproducibility and further research in the field.
\end{itemize}
\vspace{-1em}
\section{Related Work}
\textbf{Multi-view 3D Reconstruction.}
Traditional multi-view 3D reconstruction methods, as referenced in numerous studies~\cite{Kutulakos99,Fusiello00,Hartley03,Lhuillier05,Goesele06,Kazhdan06,Agarwal09,Bleyer11,Tola12,Shen13,Schonberger16}, employ geometric principles and often depend on color consistency to detect and match points across multiple images. These methods are primarily effective for static scenes with diffuse reflection. Learning-based approaches~\cite{Choy16,Qi17,Wang17,Tang18,Yao18,Chen19a,Yao19b,Chabra19,Xu20} seek to predict 3D structure from images. However, to generalize across scenes, these methods require large amounts of annotated data for training.
NeRF (Neural Radiance Fields)~\cite{Mildenhall21}, initially designed for novel view synthesis~\cite{Muller22,Verbin22,Barron21,Barron22,Barron23}, have been rapidly extended to 3D reconstruction beyond discrete points by incorporating implicit surface representations~\cite{Yariv21,Wang21b,Yu22,Wang22a,Yariv23,Wang23a,Li23} to produce smooth, continuous surfaces.
For a review of NeRF-based 3D reconstruction, please see~\cite{Gao22a}. Despite this progress, handling complex material properties in 3D reconstruction remains challenging; only a few attempts such as~\cite{Liu23b,Wang23c} aim to integrate BSDF into the neural field optimization process, thus enhancing the capability to deal with complex materials. Importantly, there is currently no extensive benchmark to evaluate such efforts in pushing the boundaries of the NeRF-based approach.

\textbf{Existing Dataset.}
Most real~\cite{Boss21,Verbin22,Kuang22,Toschi23,Liu24,Mildenhall19} and synthetic~\cite{Mildenhall21,Boss21,Verbin22} datasets designed for relighting and novel view synthesis do not encompass 3D shape, which is critical for evaluating 3D reconstruction.
While a few real datasets~\cite{Oxholm14,Aanaes16,Knapitsch17,Li20b,Kuang23,Wu23,Jampani24} feature objects composed of mixed materials with 3D annotations, the number of objects and the proportion of complex materials are relatively small. An exception is  the dataset of~\cite{Liu23b}, which contains 8 objects with smooth, shiny metal surfaces, providing higher material complexity.
Given the challenges in capturing real objects, efforts have shifted towards synthetic data creation.
ShapeNet-Intrinsics~\cite{Shi17} contains a large number of 3D shapes rendered in multiple views using a basic Phong model.
BlendedMVS~\cite{Yao20} renders textured meshes using image-based textures and standard rasterization, without modeling material-specific reflectance properties. As a result, it does not simulate per-material BRDFs or physically-based light interactions.
Objaverse~\cite{Deitke23} contains extensive 3D models collected from open-source 3D design websites, and ABO~\cite{Collins22} collects a substantial number of artist-created 3D models of household objects; however, the materials are defined by color, roughness, metallic, and transparency, which may not faithfully reflect the complex interactions of light and matter as observed in real-world materials.
Specific datasets such as~\cite{Oxholm14,Liu23b} target material properties but are restricted to less than 20 objects focusing on simple diffuse and reflective surfaces, omitting complex material behaviors such as refraction and transmission.
Here, we leverage the Objaverse 3D shapes, but incorporate laboratory-measured spectral-valued IORs within a physics-based rendering framework, creating a dataset that precisely models the interactions between light and matter to promote advances in 3D reconstruction in challenging conditions.

\vspace{-1em}
\section{Dataset Creation}
\vspace{-.5em}
Our objective is to create a dataset that enables an extensive evaluation of 3D reconstruction algorithms across a variety of materials, shapes, and lighting conditions.
Our dataset encompasses seven material types: Conductor, Dielectric, Plastic, Rough conductor, Rough dielectric, Rough plastic, and Diffuse, totaling 295 distinct material types.
It comprises 1001 scenes, where each scene features a unique shape, a unique material from our collection, and a lighting condition from 714 available High Dynamic Range Imaging (HDRI) environmental lighting options. 
To create a balanced dataset, each material type is represented in 143 scenes. In addition to multiview images from multiple camera positions, the dataset includes 3D mesh models, material annotations, camera poses, depth maps, and object masks for in-depth analysis and testing.
\vspace{-.5em}

\subsection{Physically-based Simulation}

In the real world, each material interacts with light in its own way, yielding a specific appearance.
The Bidirectional Scattering Distribution Function (\textbf{BSDF}) plays a crucial role in the rendering process by defining how light interacts with surfaces at a detailed and physically-accurate level.
It encapsulates both how light is reflected off surfaces, as described by the BRDF, and how light transmits through materials, as characterized by the Bidirectional Transmission Distribution Function (BTDF). This dual capability allows for accurate simulation of complex optical properties such as specular reflection, glossiness, and transmission.
Once the BSDF is established for a material, the color and intensity of light reflecting and transmitting at different points on the surface can be computed.
The RGB values for each pixel in the final image are derived by integrating the light from all incoming directions, factoring in the material properties defined by the BSDF, to realistically capture the appearance of materials under varied lighting conditions.

Formally, given the normal direction $\omega_n$ of a point on the object surface, the direction of the incident light $\omega_i$, and the directions of the outgoing light ($\omega_r$ for reflection and $\omega_t$ for refraction), the BSDF $f_s$ of the surface material of the object can be expressed as the sum of the BRDF $f_r$ and the BTDF $f_t$, i.e., $f_s=f_r+f_t\;.$
The BRDF and BTDF~\cite{Walter07} for microsurface reflection and refraction are defined as
\[f_r = \frac{\bF(\omega_i, \omega_h)\bG(\omega_i, \omega_r, \omega_h)\bD(\omega_h)}{4|\omega_i \cdot \omega_n||\omega_r \cdot \omega_n|},\]
\[f_t = \frac{|\omega_i \cdot \omega_h||\omega_t \cdot \omega_h|}{|\omega_i \cdot \omega_n||\omega_t \cdot \omega_n|}\frac{\eta_t^2 (1 - \bF(\omega_i, \omega_h))\bG(\omega_i, \omega_t, \omega_h)\bD(\omega_h)}{ (\eta_i (\omega_i \cdot \omega_h) + \eta_t (\omega_t \cdot \omega_h))^2},\]
where $\omega_h$ represents the intermediate vector between the $\omega_i$ and $\omega_r$.
For the BRDF, $\omega_h = sign(\omega_i \cdot \omega_n)(\omega_i+\omega_r)$; for the BTDF, $\omega_h = -(\eta_i\omega_i + \eta_t\omega_t)$. For ideal reflection and ideal refraction, $\omega_h$ is equal to the normal direction $\omega_n$. More details can be found in~\cite{Walter07}.
$\bF$ is determined using Fresnel's equation~\cite{Born13} (see below), $\bG$ is the shadow occlusion function, which we define here as Smith's shadow-masking function~\cite{Walter07}, and $\bD$ is the microfacet normal distribution function. 
Finally, $\eta_i$ and $\eta_t$ represent the IOR of the initial and transmitted media, one of which is air in our setting.


To capture the characteristics of real-world materials, our simulations employ lab-measured spectral-valued IORs. Both the angle (Snell's law) and the intensity (Fresnel's equation) of outgoing light are accurately calculated based on the IOR. By integrating these well-established physical laws with real-world IORs, we ensure an accurate physical simulation of the interactions of light and materials.


\noindent\textbf{Snell's law.}
Let $\theta_i$, $\theta_r$, $\theta_t$ represent the angles between $\omega_{i}$ and $\omega_{n}$, between $\omega_{r}$ and $\omega_{n}$,  and between $\omega_{t}$ and $\omega_{n}$, respectively.
For specular reflection, we have $\theta_i = \theta_r$.
By contrast, for transmitted light in transparent materials, the angle $\theta_{t}$ can be calculated using the IOR via Snell’s law~\cite{Born13} as
\[\theta_{t} = arcsin (\frac{\eta_{i} \sin\theta_{i}}{\eta_{t}})\;.\]

\noindent\textbf{Fresnel's equation.}
Once the angle of the outgoing light is determined, Fresnel's equation can be used to determine the precise amount of light reflected and transmitted at the interface between two media. For the reflected amount of light $\bF$, Fresnel's equation is expressed as
\begin{align*}
 \quad r_{\parallel} = \frac{\eta_t \cos \theta_i - \eta_i \cos \theta_t}{\eta_t \cos \theta_i + \eta_i \cos \theta_t}, 
 \quad r_{\perp} = \frac{\eta_i \cos \theta_i - \eta_t \cos \theta_t}{\eta_i \cos \theta_i + \eta_t \cos \theta_t}, \;\;\;
 \bF = \frac{1}{2}(r_{\parallel}^2 + r_{\perp}^2).\label{eq:fresnel}
\end{align*}
The quantity of transmitted light, such as the light passing through the dielectric material, can be calculated as $1 - \bF$, reflecting the energy conservation principles. 

\input{fig/material_physics}

\subsection{Spectral-valued Index of Refraction}

In this work, we employ spectral-valued IORs~\cite{Jakob19} rather than conventional single-value IORs~\cite{Chen23b} to more accurately simulate the real-world spectral nature of material IORs, capturing the complex interactions of light with materials.
These spectral-valued IORs are derived from laboratory measurements. They cover a variety of materials, including gold, silver, copper, and rhodium, used in manufacturing and coatings, along with commercial glass and plastics. 

Our simulations are thus adjusted to the material at hand to account for the variability in IOR across different wavelengths.
One advantage of spectral-valued IORs is that they simulate light and material interactions more accurately than single-value IORs~\cite{Chen23b}. 
Traditionally, simulations use single-valued IORs to mimic the color of reflected light on metal surfaces, complemented by a preset color to represent the metal's inherent color. However, this approach fails to capture the true physical processes, leading to inaccurate results.
In contrast, spectral-valued IORs vary with wavelength—lower at shorter wavelengths (closer to blue light) and higher at longer wavelengths (toward the red spectrum). According to Fresnel's equations, this variability means that, for example, gold absorbs more light at shorter wavelengths and reflects more at longer wavelengths, giving it its characteristic golden-yellow appearance. Thus, employing spectral-valued IORs significantly enhances the accuracy of simulations by more faithfully reproducing light-material interactions.

The second advantage of using spectral-valued IORs to accurately simulate physical processes is that, integrated with fundamental physical models, they enable algorithms to extract more detailed information from phenomena such as reflections and highlights. 
Using spectral-valued IORs in simulations replicates the real physical processes of phenomena like reflection and highlights. 
Integrating these physical processes into the algorithm's design could potentially enhance the algorithm's ability to manage complex optical phenomena associated with complex materials.
An example of this is NeRO~\cite{Liu23b}, one of the current leading algorithms for processing reflective surfaces, which accurately models both the incident ($\omega_i$) and reflected ($\omega_o$) light to extract the information necessary to reconstruct reflective surfaces. Specifically, NeRO models specular light as
\[
c_{\text{specular}} \approx \underbrace{\int_{\Omega} L(\omega_i)D(\rho, \omega_r)d\omega_i}_{L_{\text{specular}}} \cdot \underbrace{\int_{\Omega} \frac{DFG}{4(\omega_o \cdot n)} d\omega_i}_{M_{\text{specular}}},
\]
where $L_{\text{specular}}$ is the integral of light on the normal distribution function $D(\rho, \omega_r)\in[0, 1]$, $\rho$ is the roughness, $\omega_o$ is the viewing direction, $M_{\text{specular}}$ denotes the integral of the BRDF, $F$ is the Fresnel term, and $G$ is the geometry term. More details can be found in~\cite{Liu23b}.
As shown in the results of~\cref{tab:neus}, NeRO significantly outperforms the algorithms that model only the incident light.

\subsection{Microfacet Model}
Let us now describe the microfacet model we used to create our dataset.
Given the BSDF, we exploit a microfacet model to realistically model the microscopic surface details that affect light scattering. Specifically, we adapt the model for smooth and rough surfaces by selecting appropriate microfacet distributions.
For materials such as conductors, dielectrics, and plastics with smooth surfaces, we employ the Dirac delta distribution $D(\omega_h) = \delta(\omega_h - \omega_{n})$, which is non-zero only at the surface normal orientation $\omega_{n}$. This choice ensures sharp and precise reflections and refractions, as evidenced in~\cref{fig:material}, effectively simulating the realistic appearance of polished metals, glass, and plastic.

For materials with rough surfaces, including rough conductors, dielectrics, and plastics, we select the Trowbridge-Reitz (GGX) distribution to simulate glossy specular reflections and diffuse refractions.
This distribution, characterized by its heavy tail influenced by the roughness parameter $\alpha$, more accurately models the complex scattering of light on rough surfaces, enhancing the accuracy of physical simulations, as illustrated in~\cref{fig:material}.
In this case, we have
\[D(\omega_h)=\frac{\alpha^2}{\pi\left(\cos^2\theta_h(\alpha^2-1)+1\right)^2}\;,\]
where $\alpha$ is the surface roughness parameter, and $\theta_h$ is the angle between $\omega_{h}$ and $\omega_{n}$.


\begin{figure*}[tb]
    \setlength{\belowcaptionskip}{-5pt}
    \centering
    \includegraphics[width=.19\textwidth]{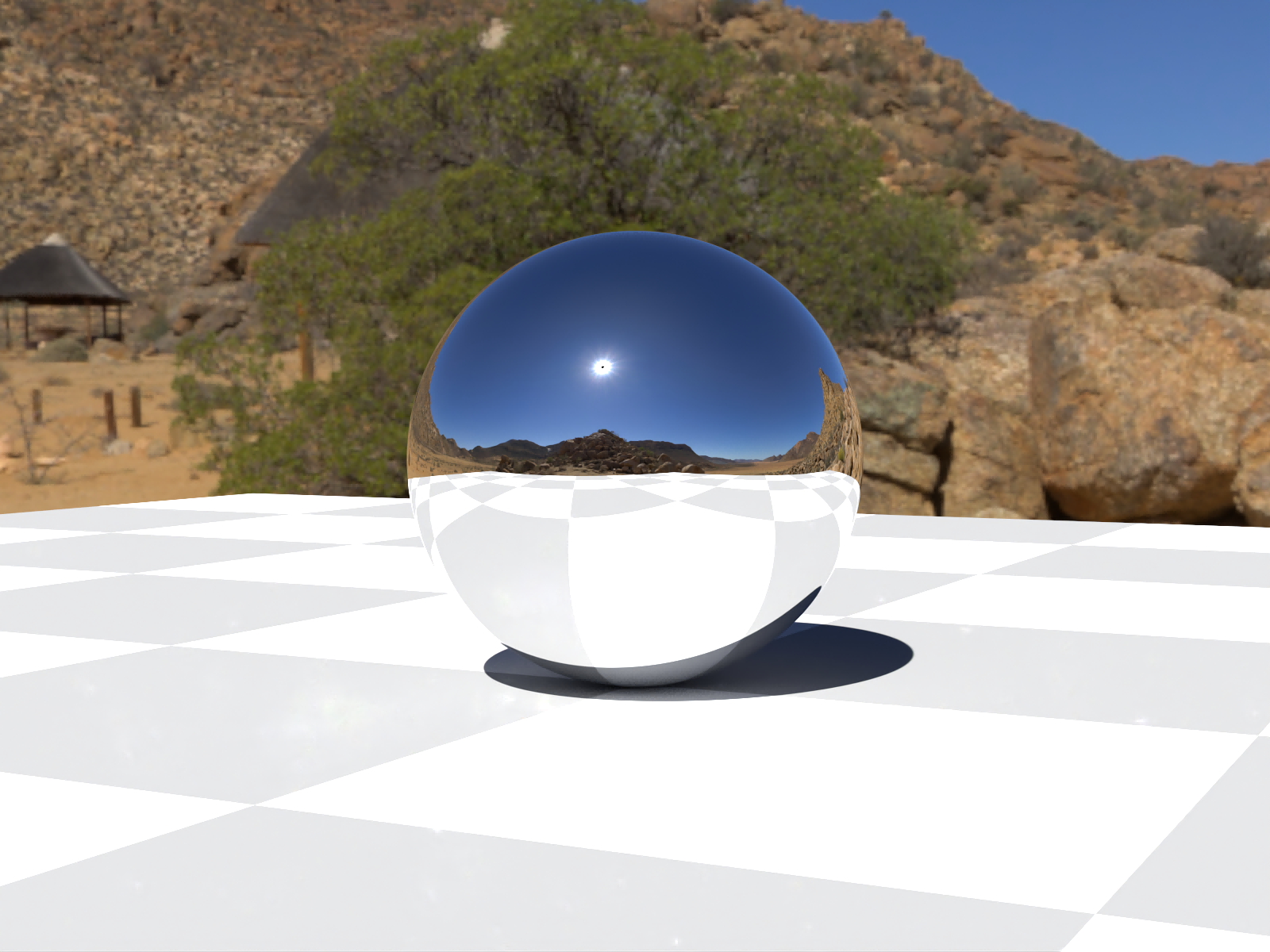}
    \includegraphics[width=.19\textwidth]{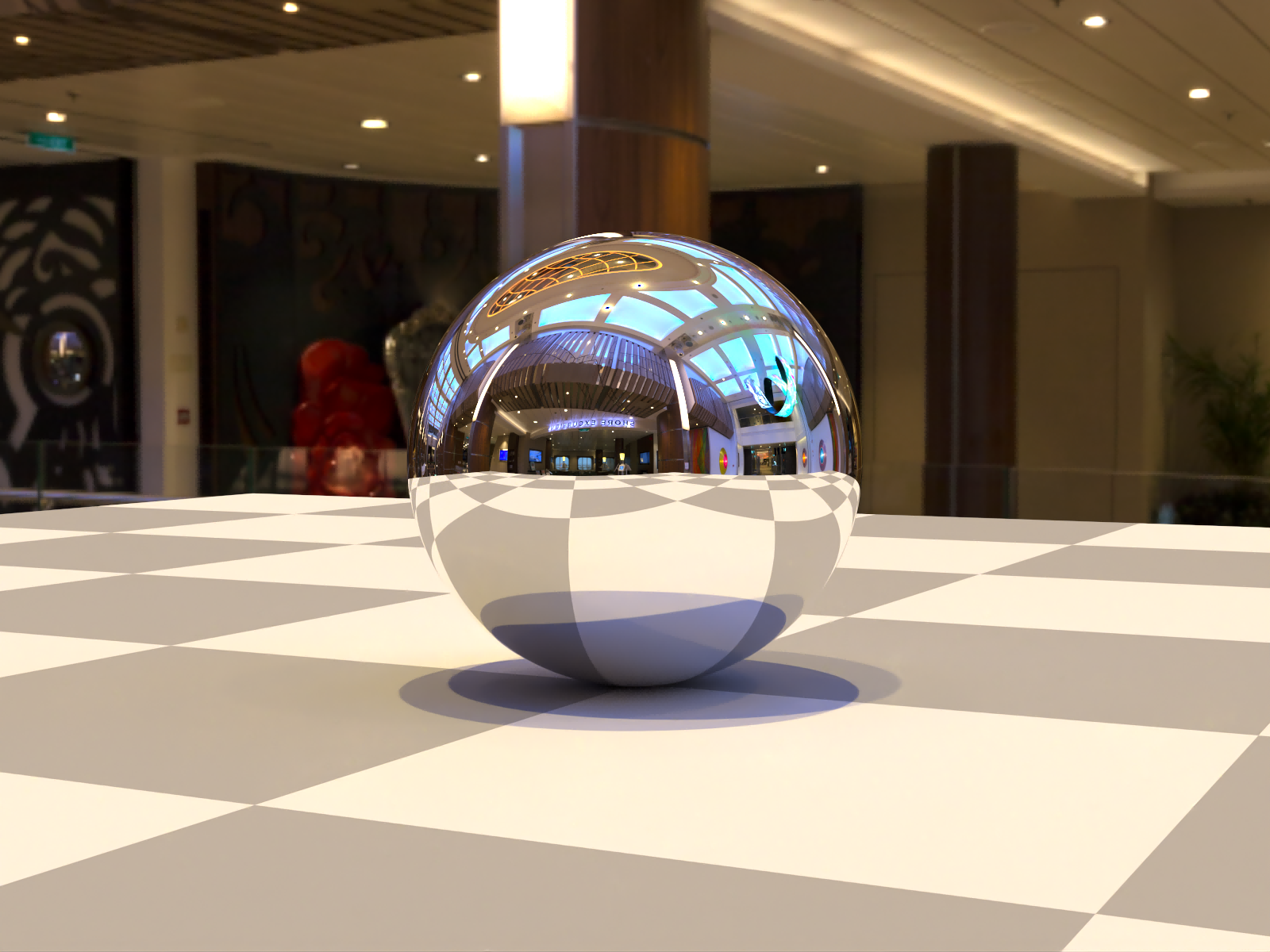}
    \includegraphics[width=.19\textwidth]{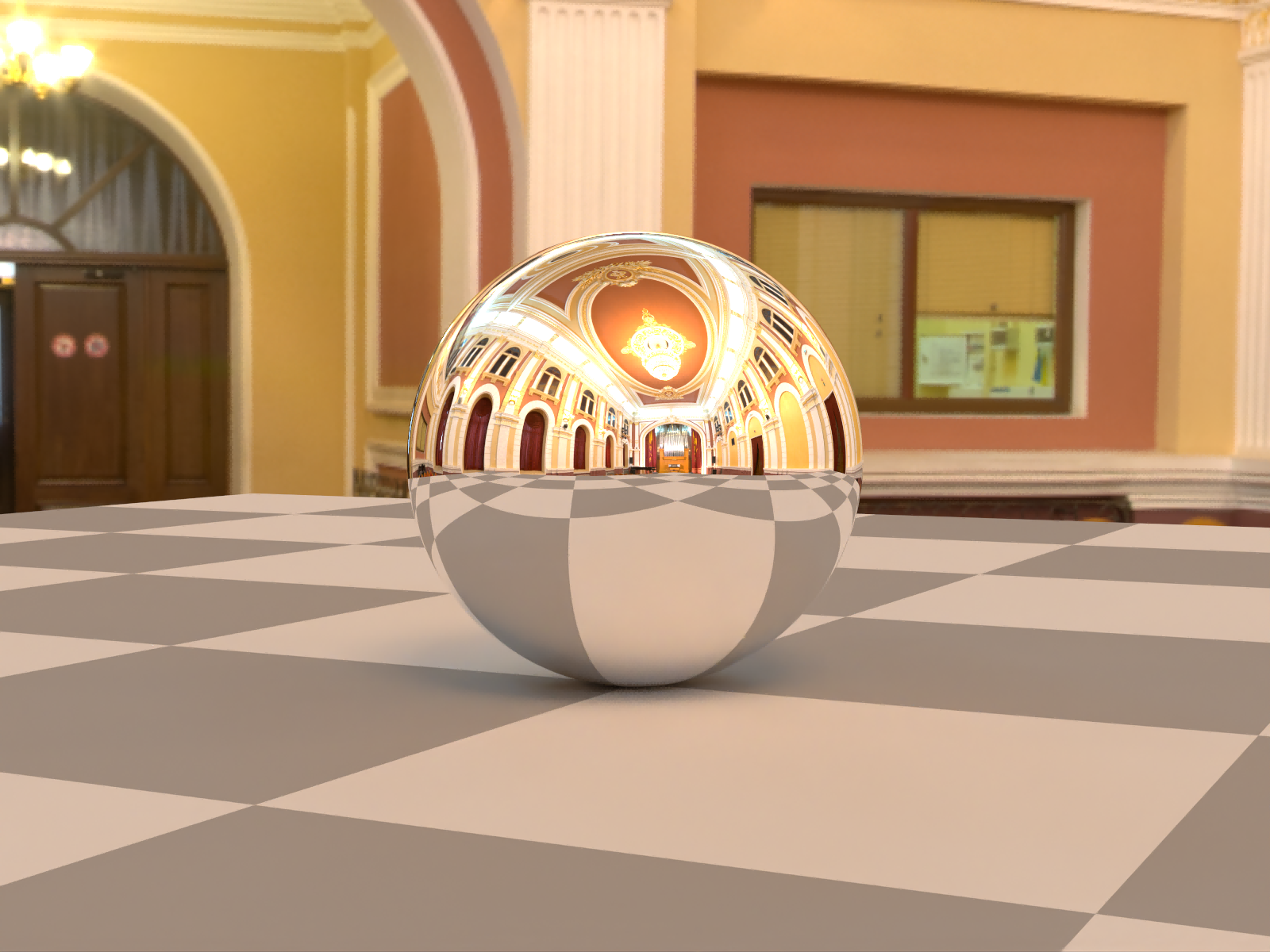}
    \includegraphics[width=.19\textwidth]{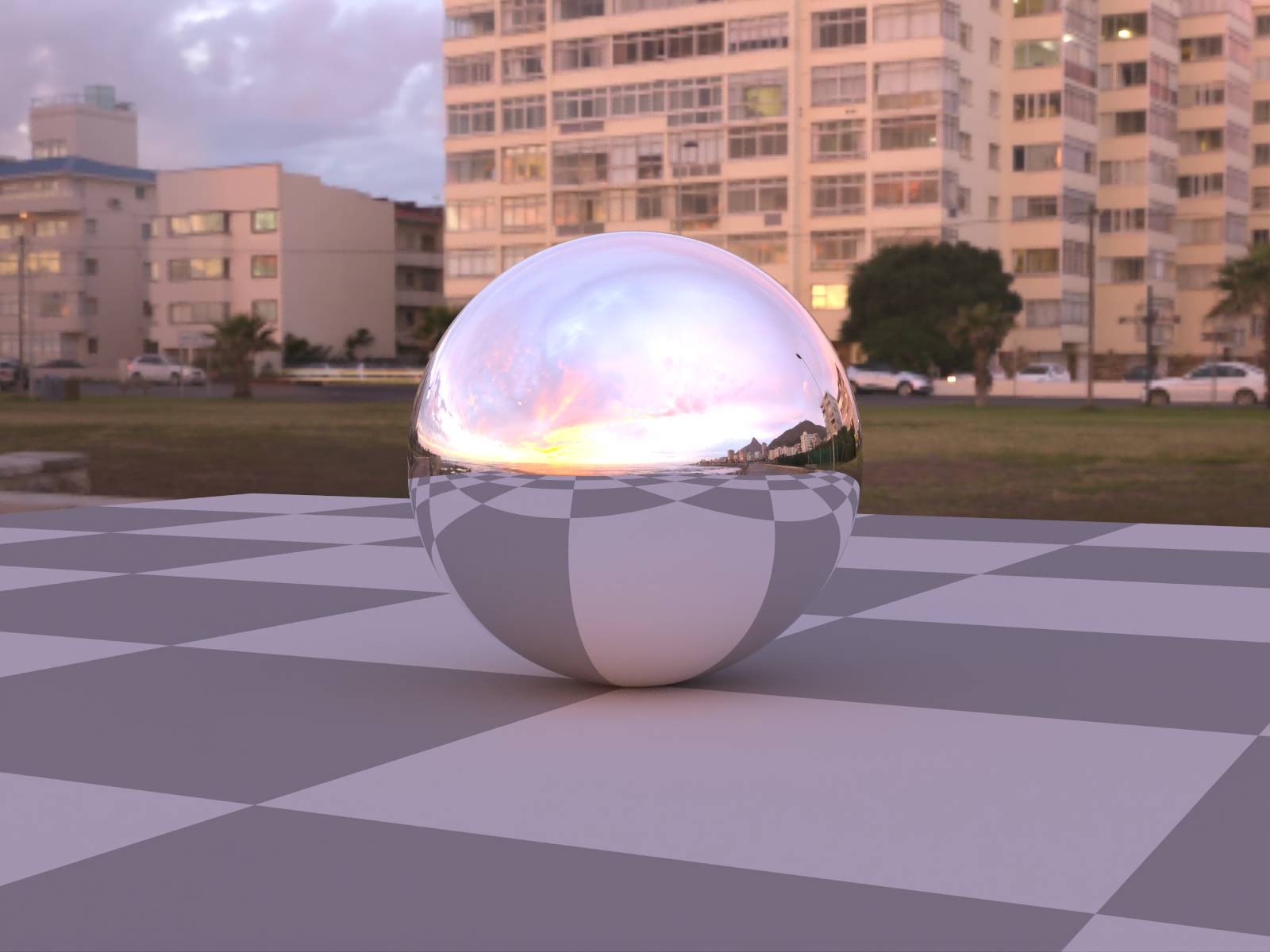}
    \includegraphics[width=.19\textwidth]{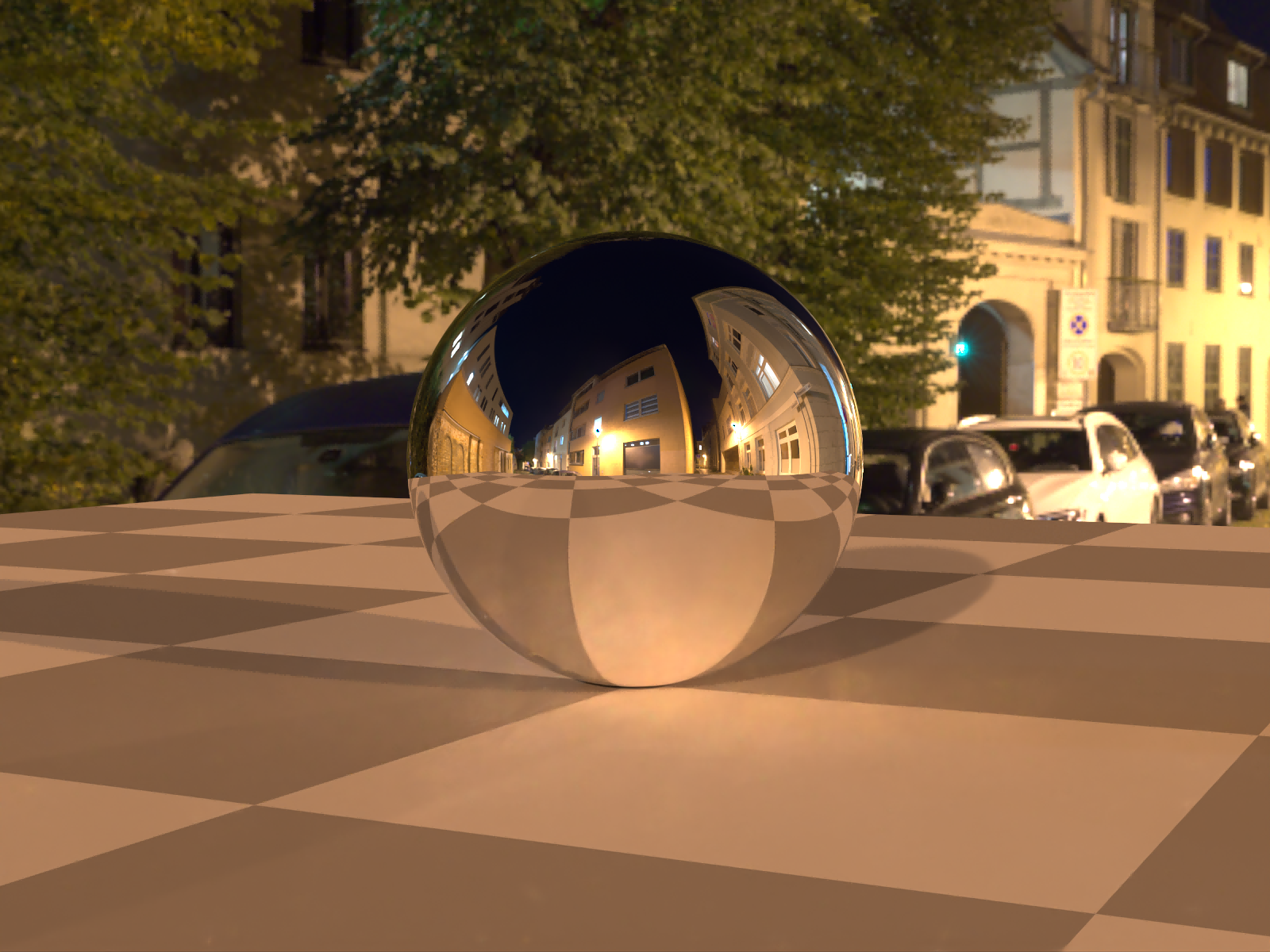}
    \caption{\textbf{Illustration of selected HDRI environment maps.}}
    \label{fig:emitter}
\end{figure*}
\subsection{Scene Illumination and Cameras}

To obtain a high diversity of lighting conditions, we utilized 714 HDRI maps from both indoor and outdoor environments. These HDRI maps capture the full spectrum of light intensities found in real-world environments, thereby increasing the dataset complexity in lighting, reflections, and refractions, as shown in~\cref{fig:emitter}. This diversity is crucial for extensively evaluating an algorithm's performance across varied lighting scenarios.

To eliminate scale bias, we standardized the object sizes within a unit sphere and sampled camera positions using a Fibonacci grid on the upper hemisphere to ensure uniform distribution. We divided these positions into 50 training and 40 testing viewpoints. 
All data, including high-resolution images (1600x1200 pixels), camera positions, depth, 3D models, and object masks, are stored in the standard COLMAP format.

\subsection{Shapes and Meshes}

Our dataset contains 1,001 shapes, showcasing a wide diversity in terms of geometry. These shapes were selected from the Objaverse-1 dataset~\cite{Deitke23} based on three main criteria. 
First, we chose only watertight objects with volume; for instance, flat constructions, such as a rose made from pure planes, were excluded because they do not realistically represent solid objects. 
\begin{wrapfigure}{r}{0.5\linewidth}
    \centering
    \setlength{\belowcaptionskip}{-20pt}
    \includegraphics[width=\linewidth]{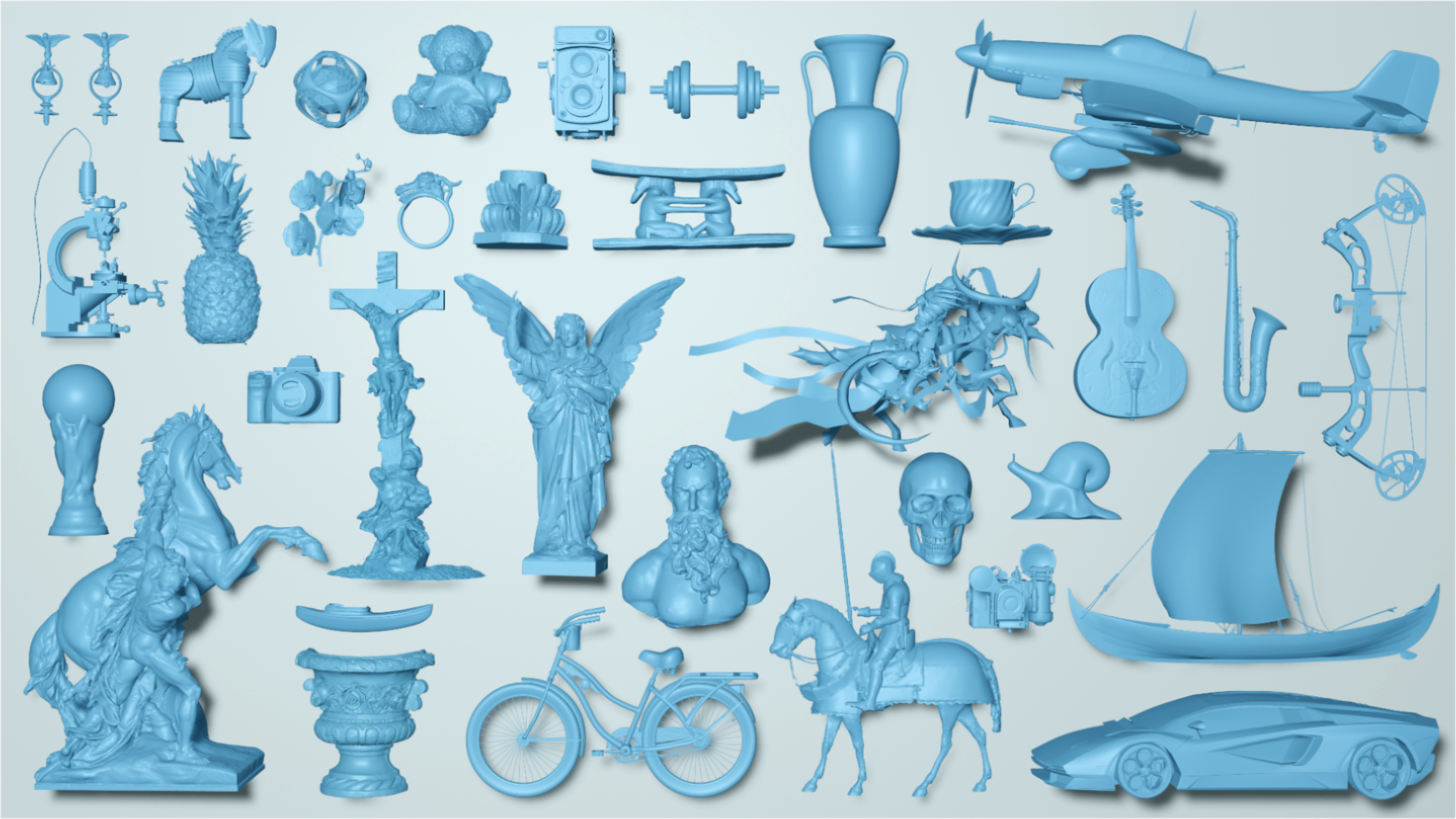}
    \caption{\textbf{Selection of objects.}}
    \label{fig:shape}
\end{wrapfigure}Second, our selection focused on objects with real-world counterparts, such as sculptures and vehicles, and includes some animated characters. Notable examples include ``The Kiss'' from Musée Rodin, whose mesh model was constructed from real-world scans. Third, we aimed to ensure diversity and complexity in our choices, ranging from simple items such as guitars to complex structures such as fountain statues, which feature a variety of human poses and intricate facial expressions. Some of the shapes are depicted in~\cref{fig:shape}.


\begin{table*}[b]
\centering
\scriptsize
\setlength{\tabcolsep}{3.3pt}
\setlength{\belowcaptionskip}{-15pt}
\begin{tabular}{p{1.cm}|c|cccccccc}
\toprule
\multirow{2}{*}{\parbox{1.cm}{\centering Metric}} & \multirow{2}{*}{Method} & \multicolumn{7}{c}{\textbf{Material Type}} \\
               &           & Conductor  & Dielectric & Plastic  & Rough Conductor & Rough Dielectric    & Rough Plastic & Diffuse\\
\midrule
\multirow{5}{*}{\parbox{1.cm}{\centering Chamfer Distance$\downarrow$}}  
                        & Instant-NeuS~\cite{Guo22}  & \cellcolor{top3}1.2554    & \cellcolor{top2}1.1912    & 0.7023                  & \cellcolor{top3}0.9824  & \cellcolor{top2}1.1634  & 0.6417                  & 0.6425\\
                        & NeuS2~\cite{Wang23a}       & 1.3016                    & 1.2734                    & \cellcolor{top3}0.6823  & 1.0192                  & \cellcolor{top3}1.2553  & \cellcolor{top3}0.6372  & 0.5998\\
                        & 2DGS~\cite{Huang24}        & 2.1009                    & 1.9677                    & 0.8163                  & 1.6443                  & 1.5882                  & 0.6385                  & \cellcolor{top3}0.5549 \\
                        & PGSR~\cite{Chen24}                       & \cellcolor{top2}1.1381    & \cellcolor{top1}1.1086    & \cellcolor{top1}0.5389  & \cellcolor{top2}0.9146  & \cellcolor{top1}0.8198  & \cellcolor{top1}0.4391  & \cellcolor{top1}0.4512 \\
                        & NeRO$\star$~\cite{Liu23b}         & \cellcolor{top1}0.9796    & 1.8357                    & \cellcolor{top2}0.6413  & \cellcolor{top1}0.8320  & 1.6504                  & \cellcolor{top2}0.4907  & \cellcolor{top2}0.5105  \\
                        & NeRRF$\star$~\cite{Chen23b}       & 1.2667                    & \cellcolor{top3}1.2257    & 1.2029                  & 1.2452                  & 1.2894                  & 1.2127                  & 1.2824  \\
\bottomrule
\end{tabular}
\caption{\textbf{Comparison of SOTA 3D Reconstruction Methods.} We report the Chamfer distance $\times 10^{-2}$. A $\star$ denotes specialized methods.}
\label{tab:neus}
\end{table*}

\section{Benchmarking and Ablation Study}
Our dataset is characterized by a wide diversity of material types, geometric shapes, and lighting conditions, making it ideal for benchmarking purposes. We have selected two representative tasks for evaluation: 3D shape reconstruction and novel view synthesis. In addition, we perform an ablation study to analyze the impact of specific variables. 
Here, we highlight our key findings; for further results and visualizations, please refer to the supplementary material.

\begin{figure*}[tb]
    \setlength{\belowcaptionskip}{-.5cm}
    \centering
    \renewcommand{\arraystretch}{0.8}
    \setlength{\tabcolsep}{2pt}
    \begin{tabular}{cccc}
        Conductor & Dielectric  & Plastic &  Diffuse  \\
        \includegraphics[width=.24\textwidth]{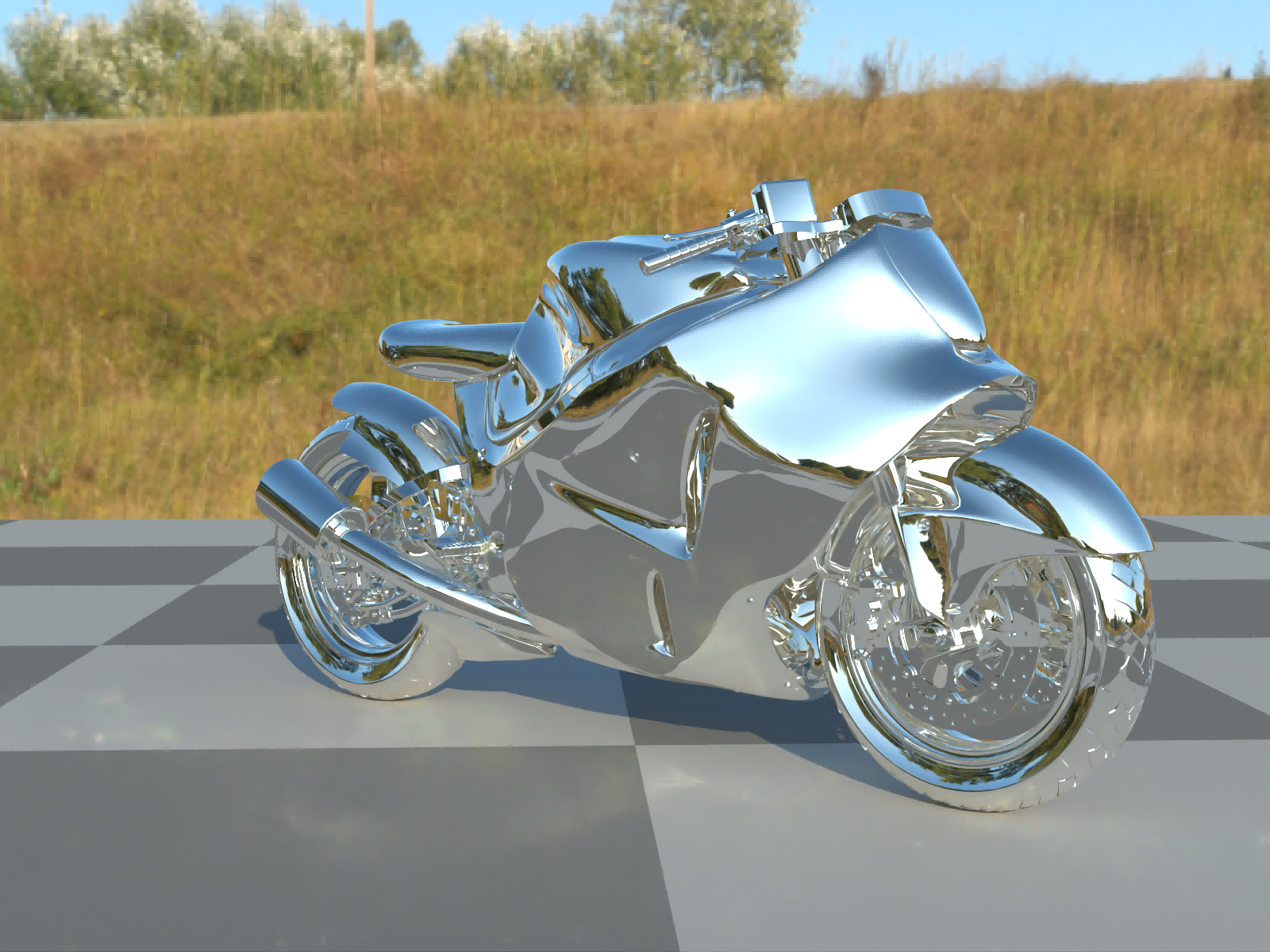} & 
        \includegraphics[width=.24\textwidth]{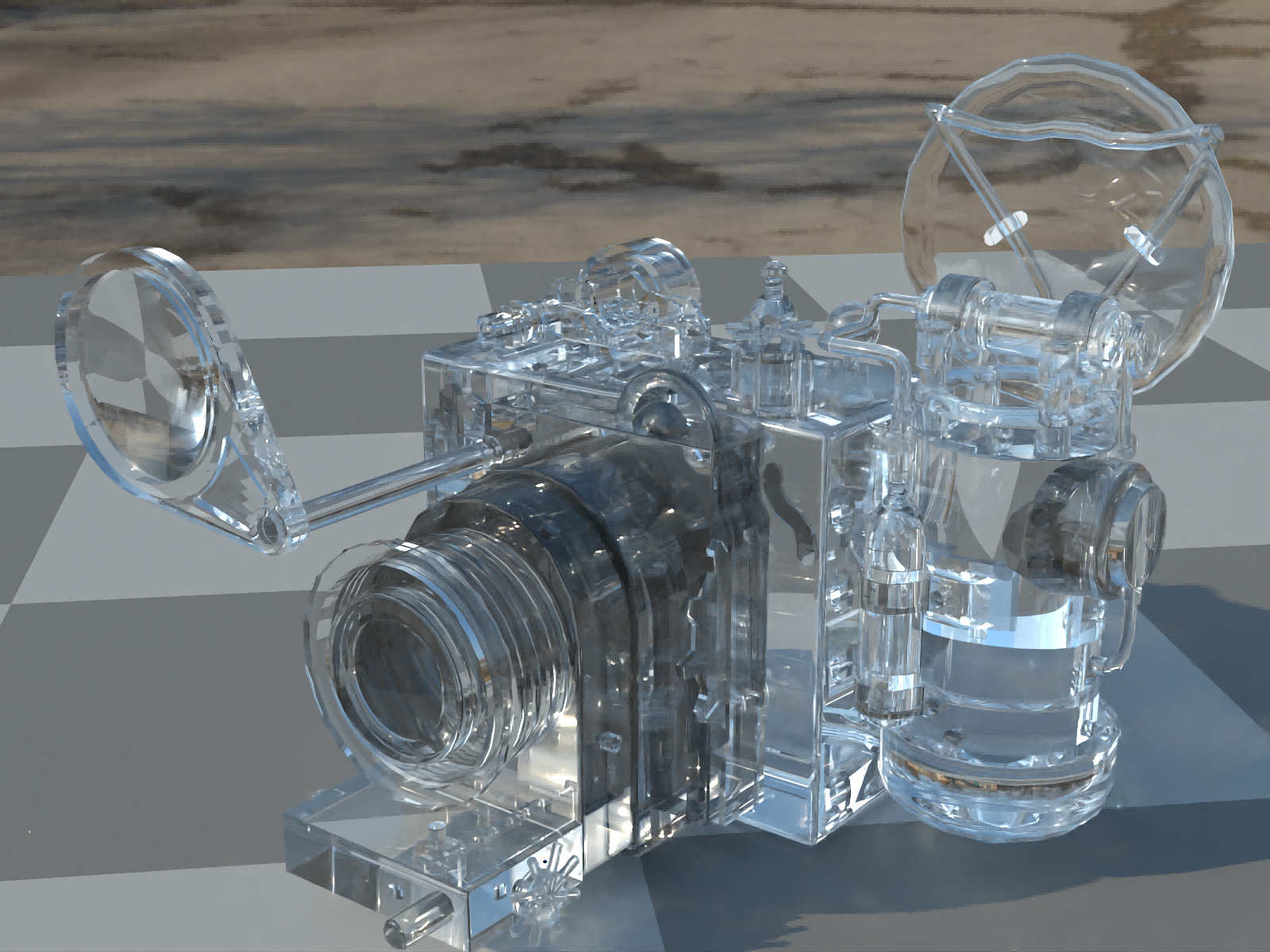} &
        \includegraphics[width=.24\textwidth]{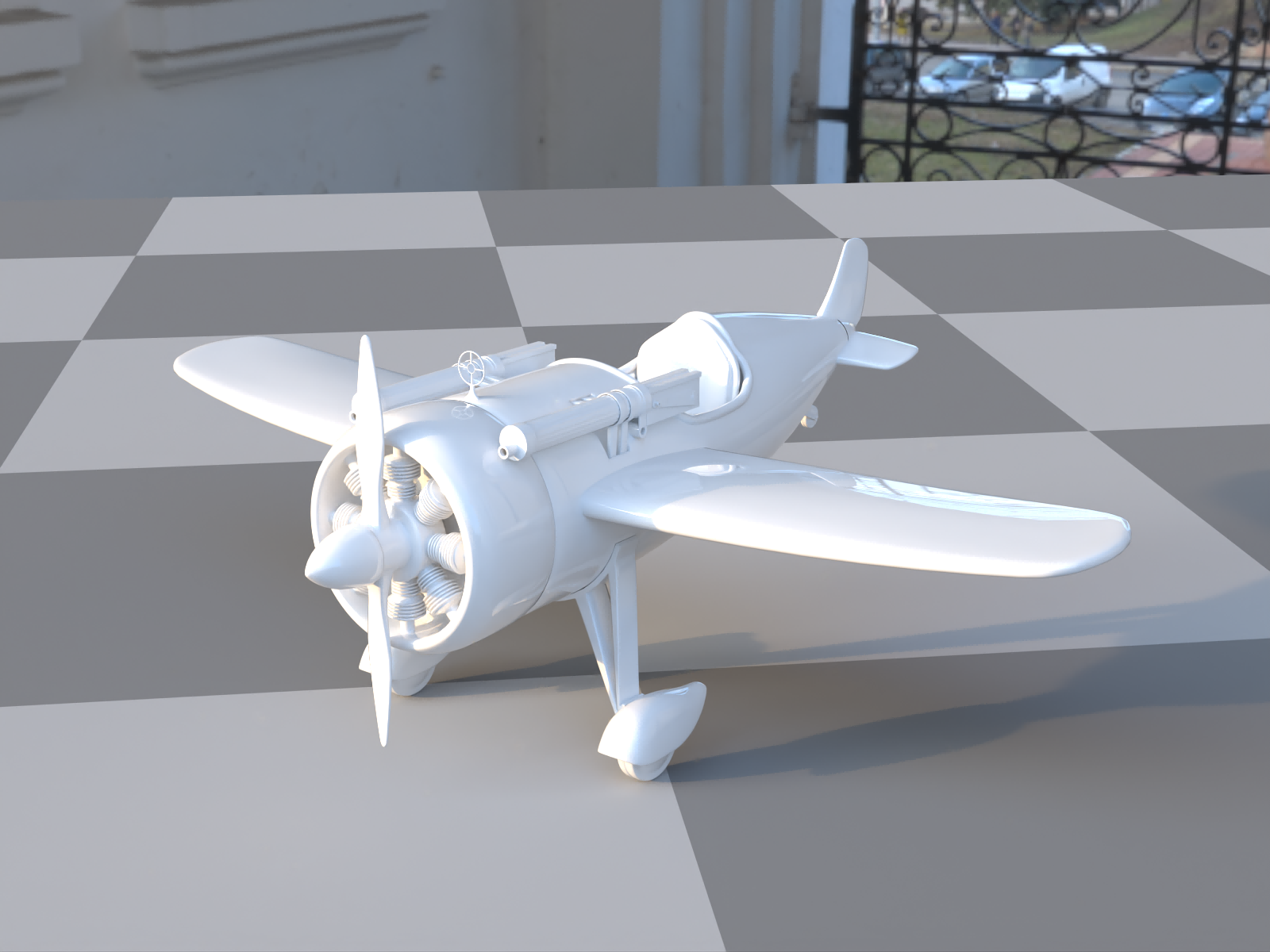} &
        \includegraphics[width=.24\textwidth]{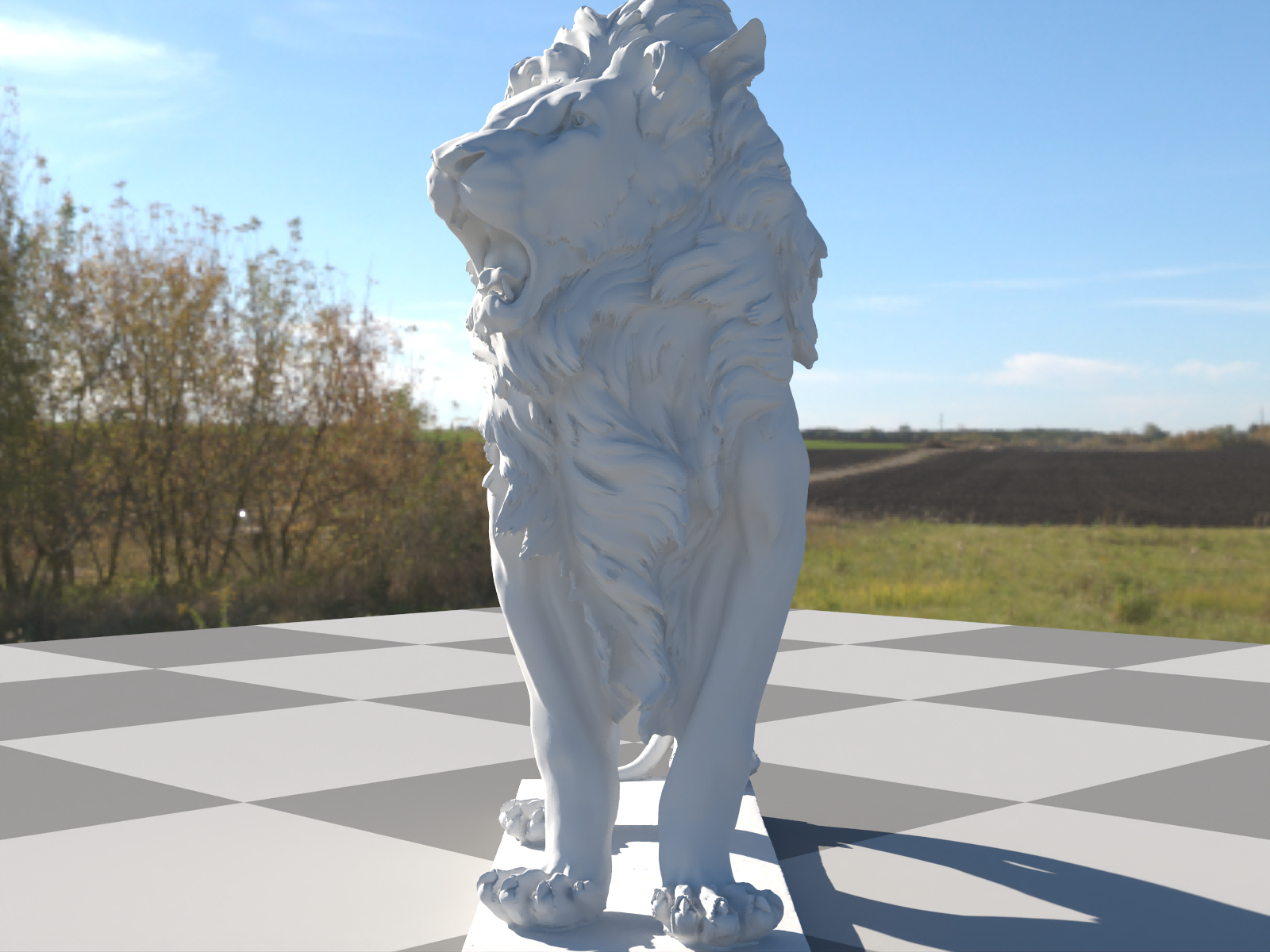} \\
        
        \includegraphics[width=.24\textwidth]{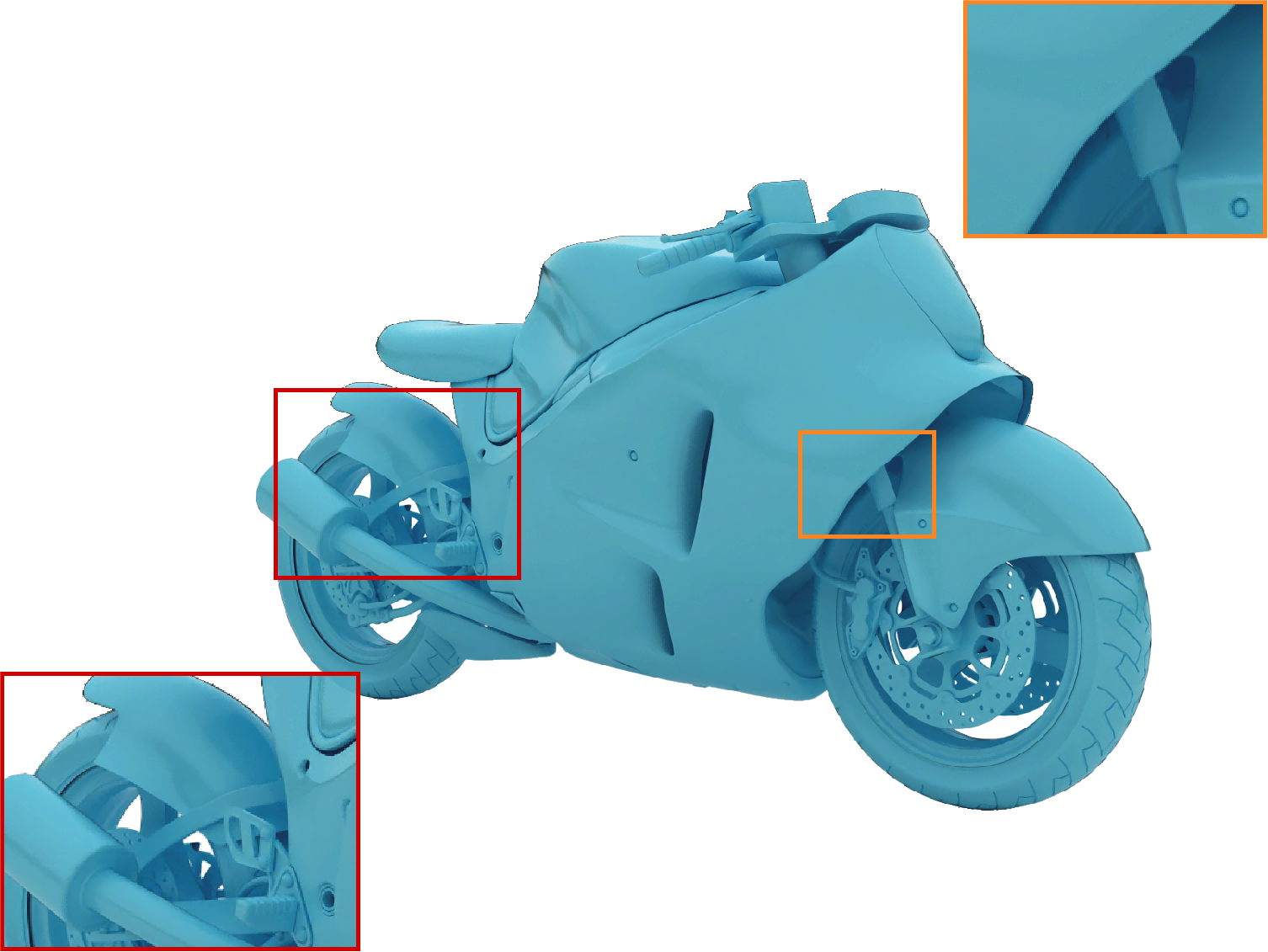} &
        \includegraphics[width=.24\textwidth]{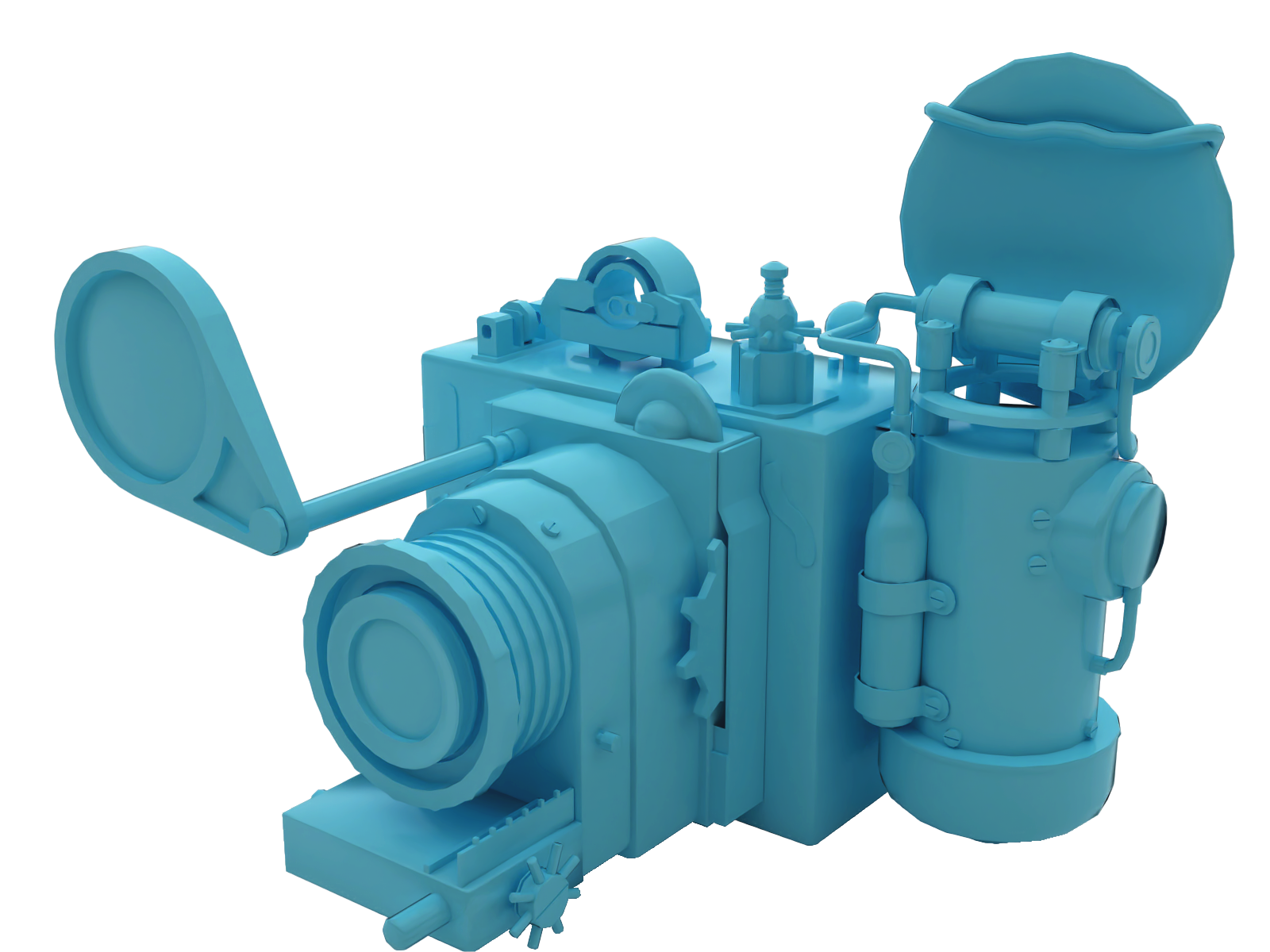} &
        \includegraphics[width=.24\textwidth]{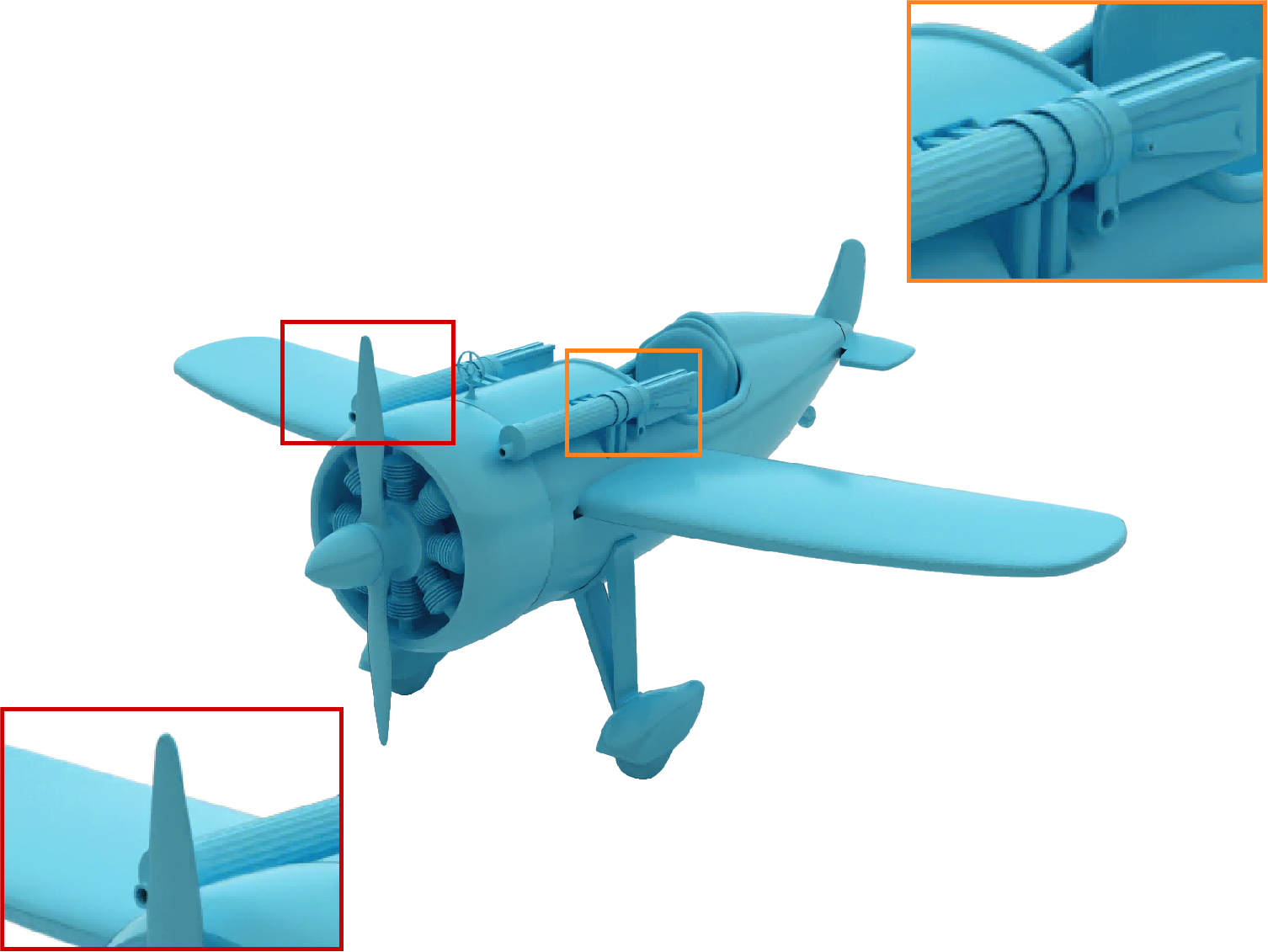} &
        \includegraphics[width=.24\textwidth]{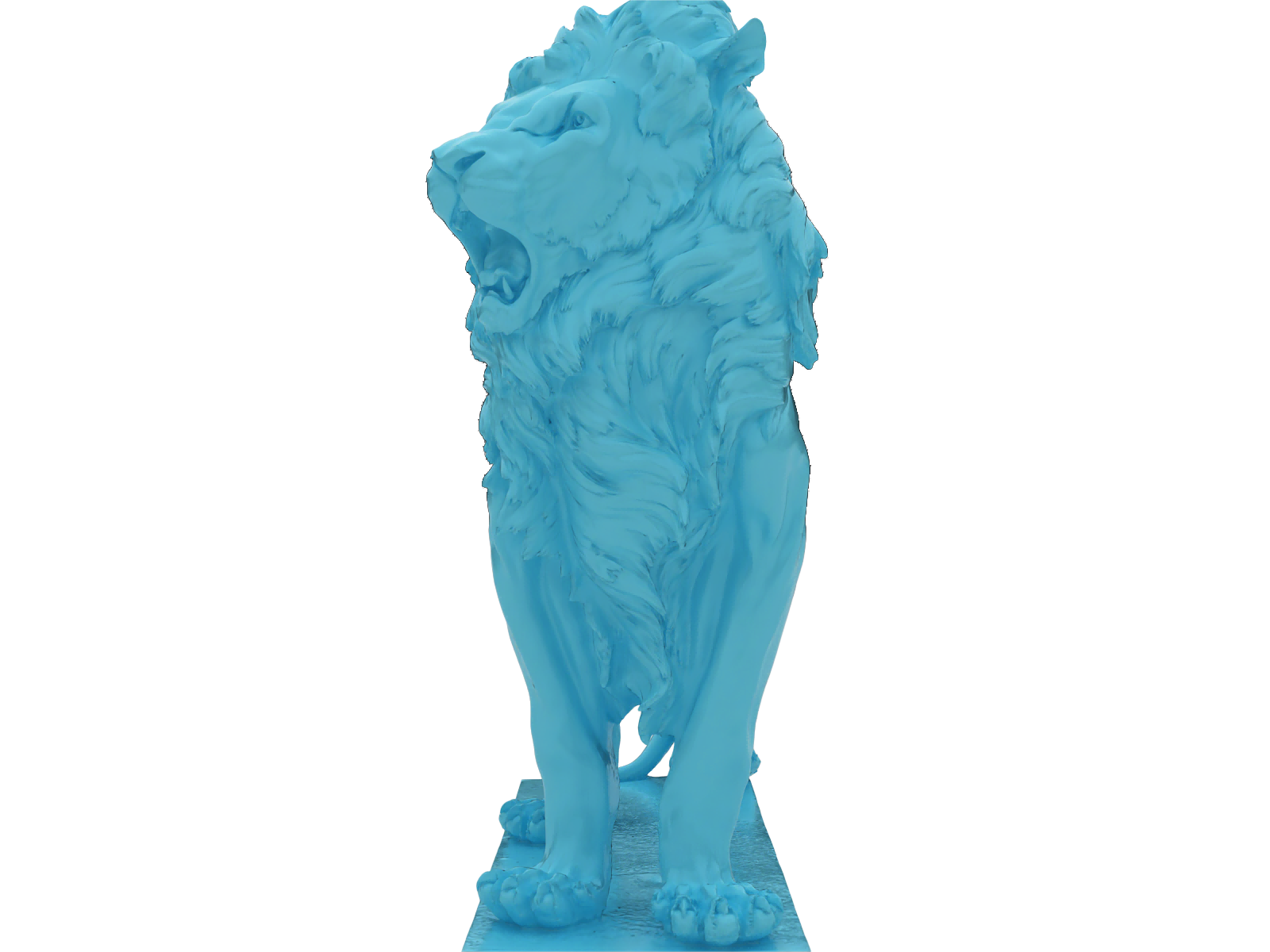} \\
        
        \includegraphics[width=.24\textwidth]{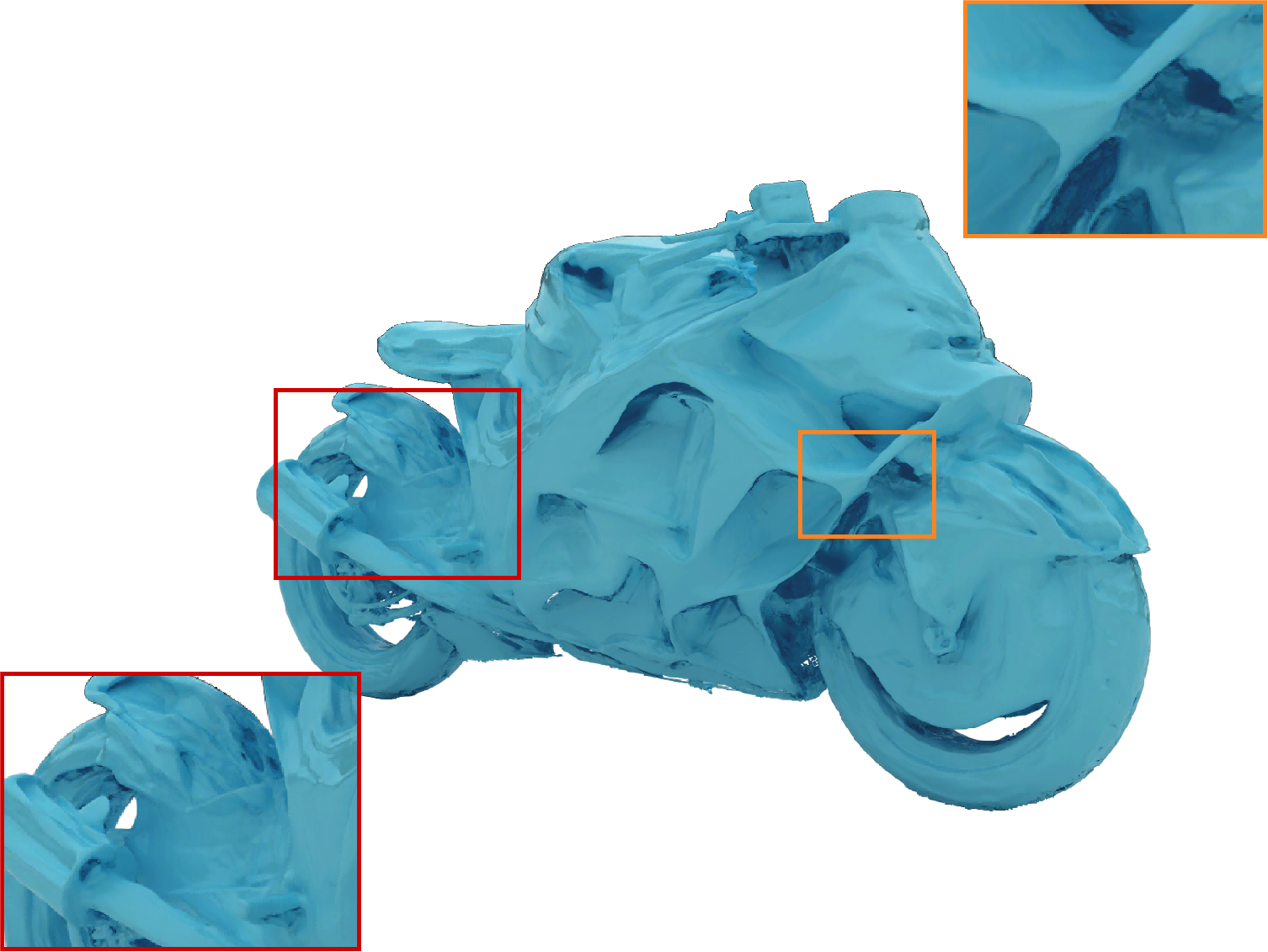}&
        \includegraphics[width=.24\textwidth]{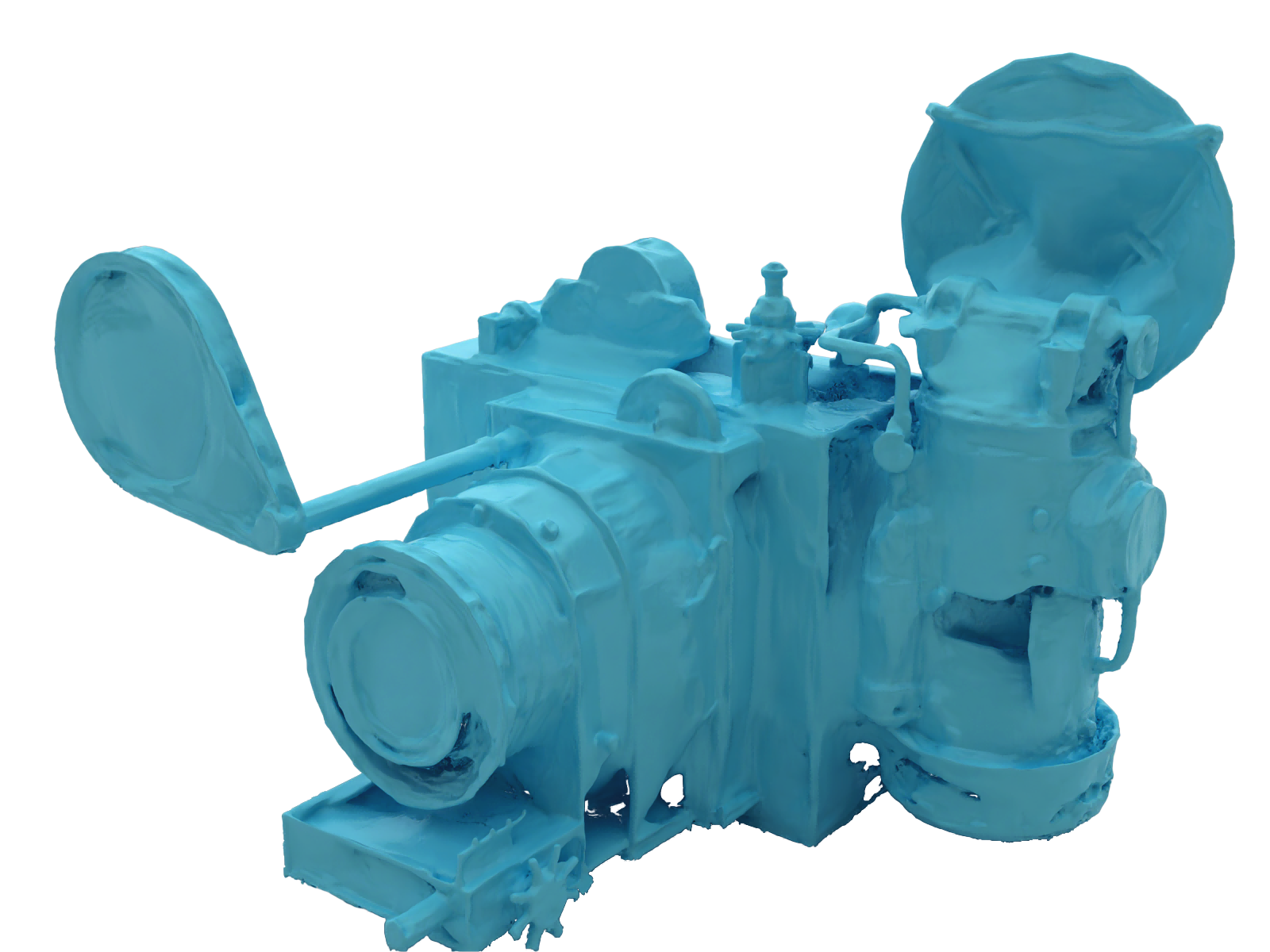}&
        \includegraphics[width=.24\textwidth]{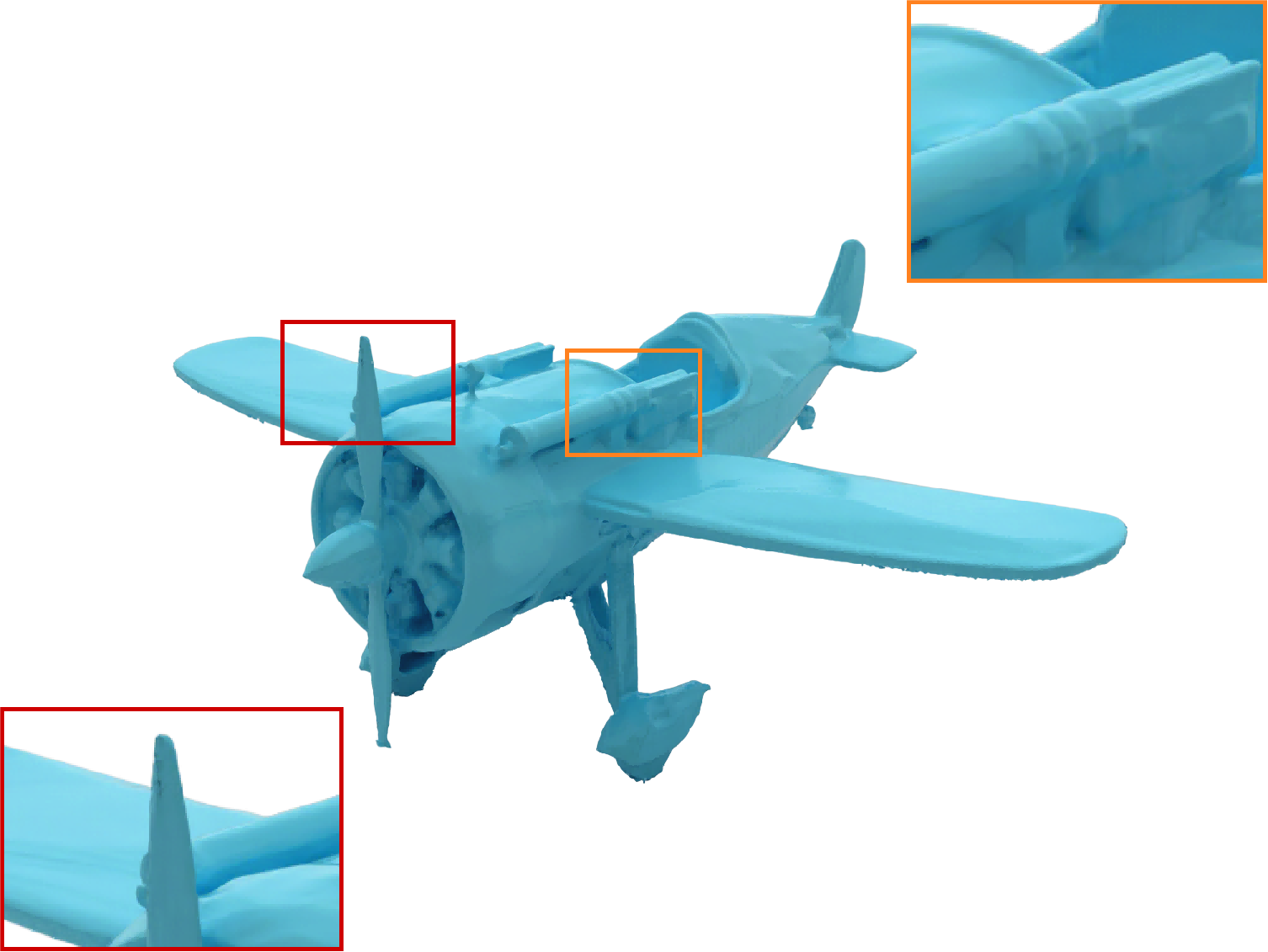} &
        \includegraphics[width=.24\textwidth]{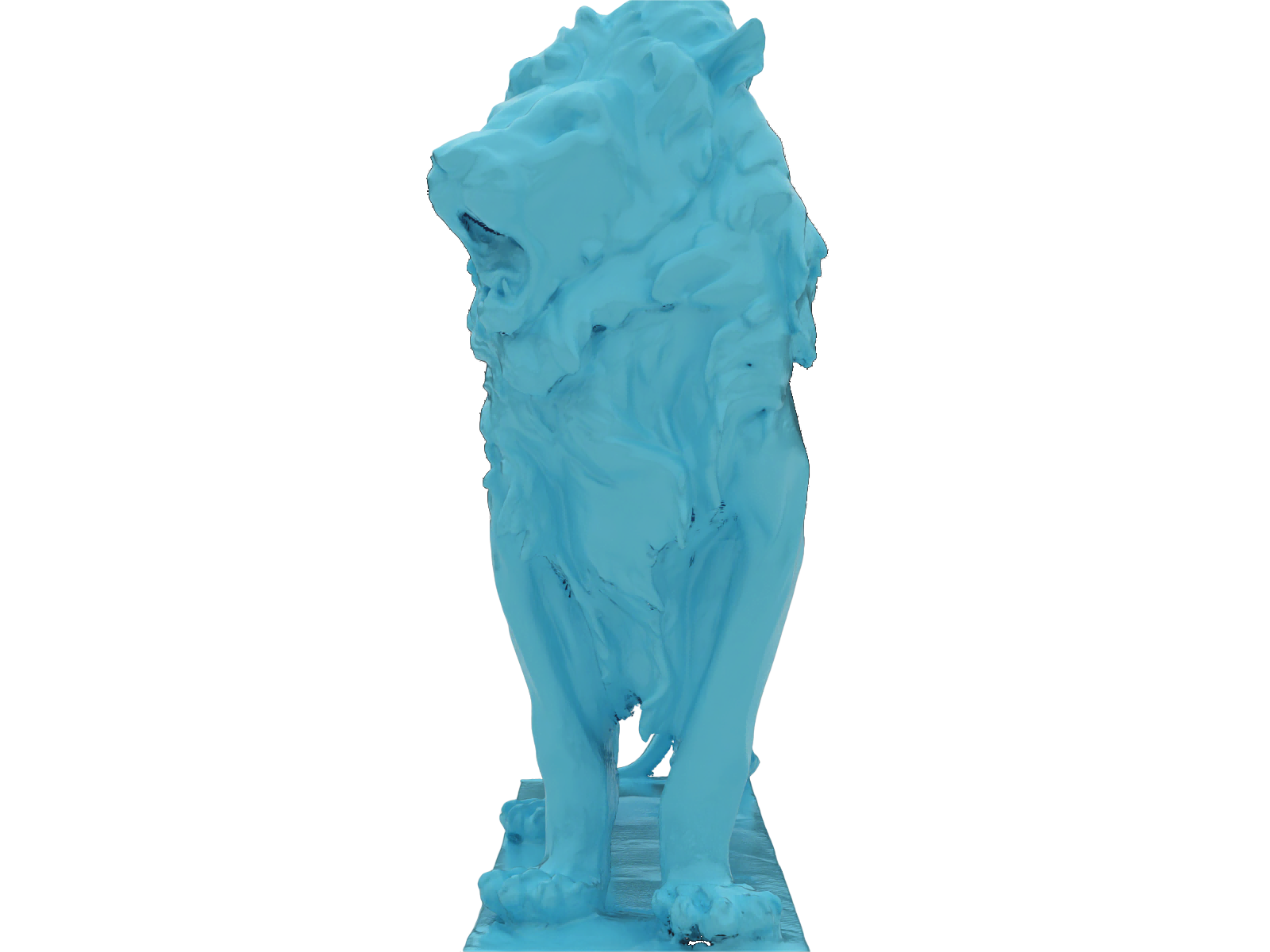} \\
        
        \includegraphics[width=.24\textwidth]{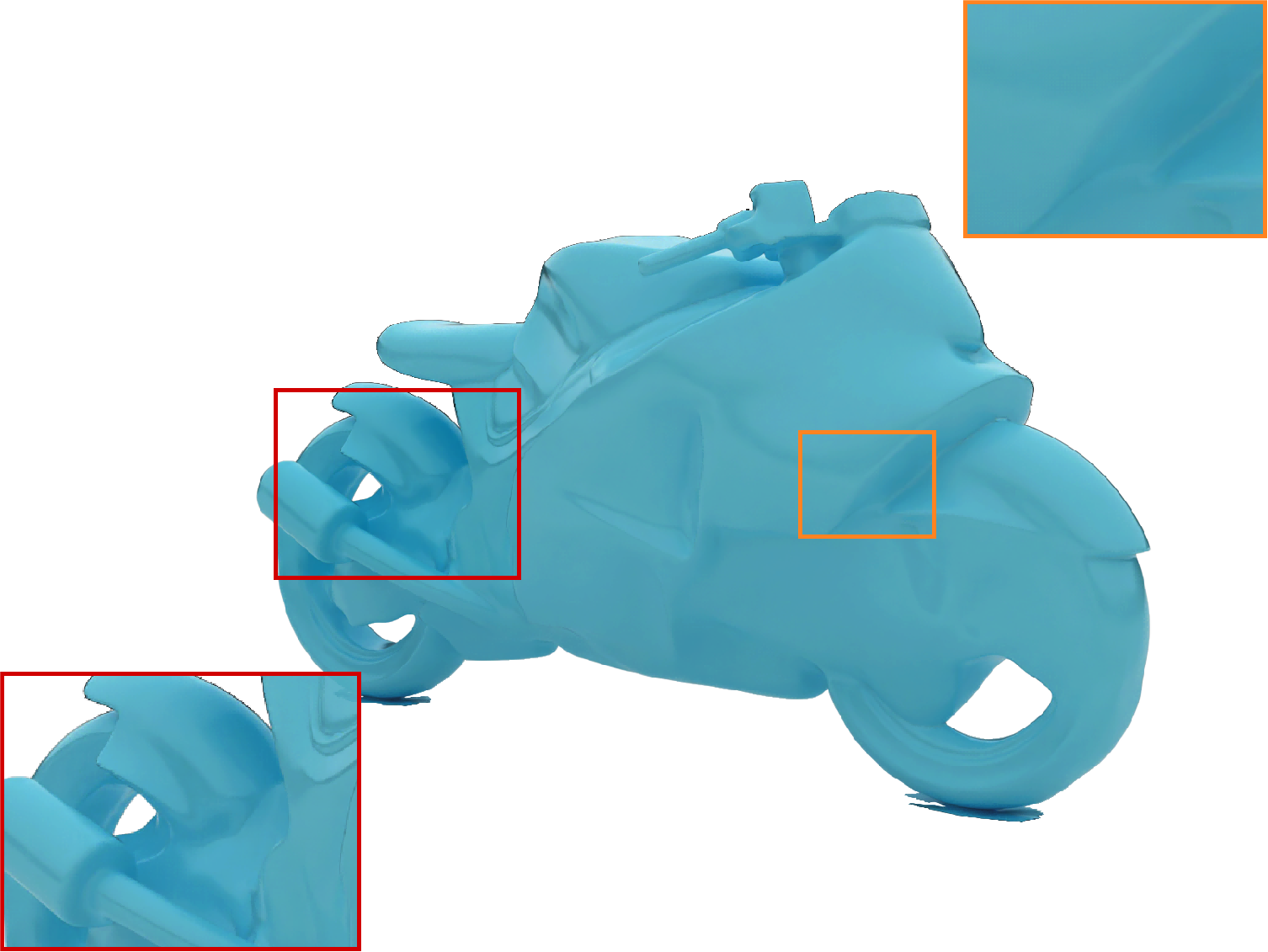} &
        \includegraphics[width=.24\textwidth]{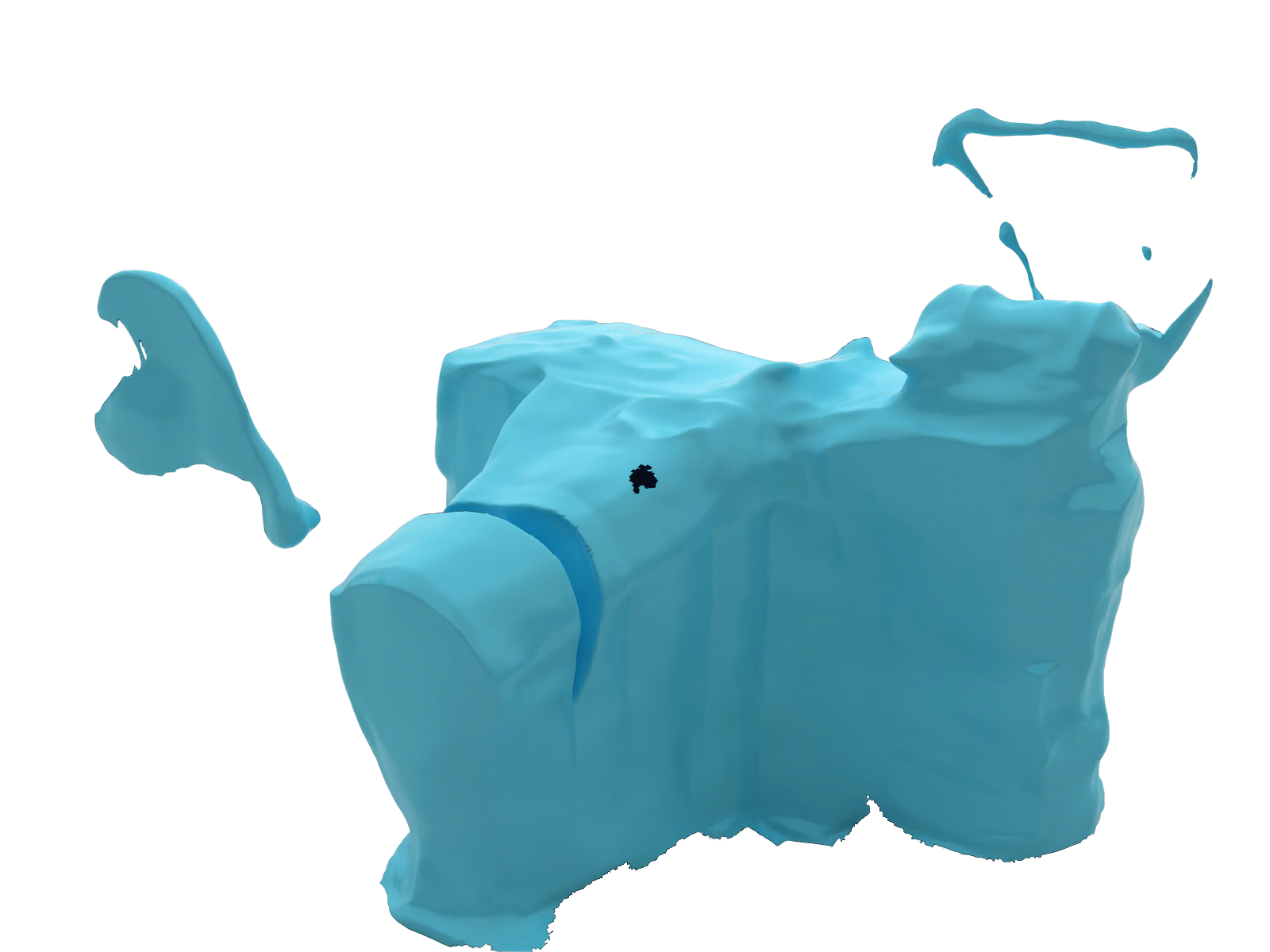}&
        \includegraphics[width=.24\textwidth]{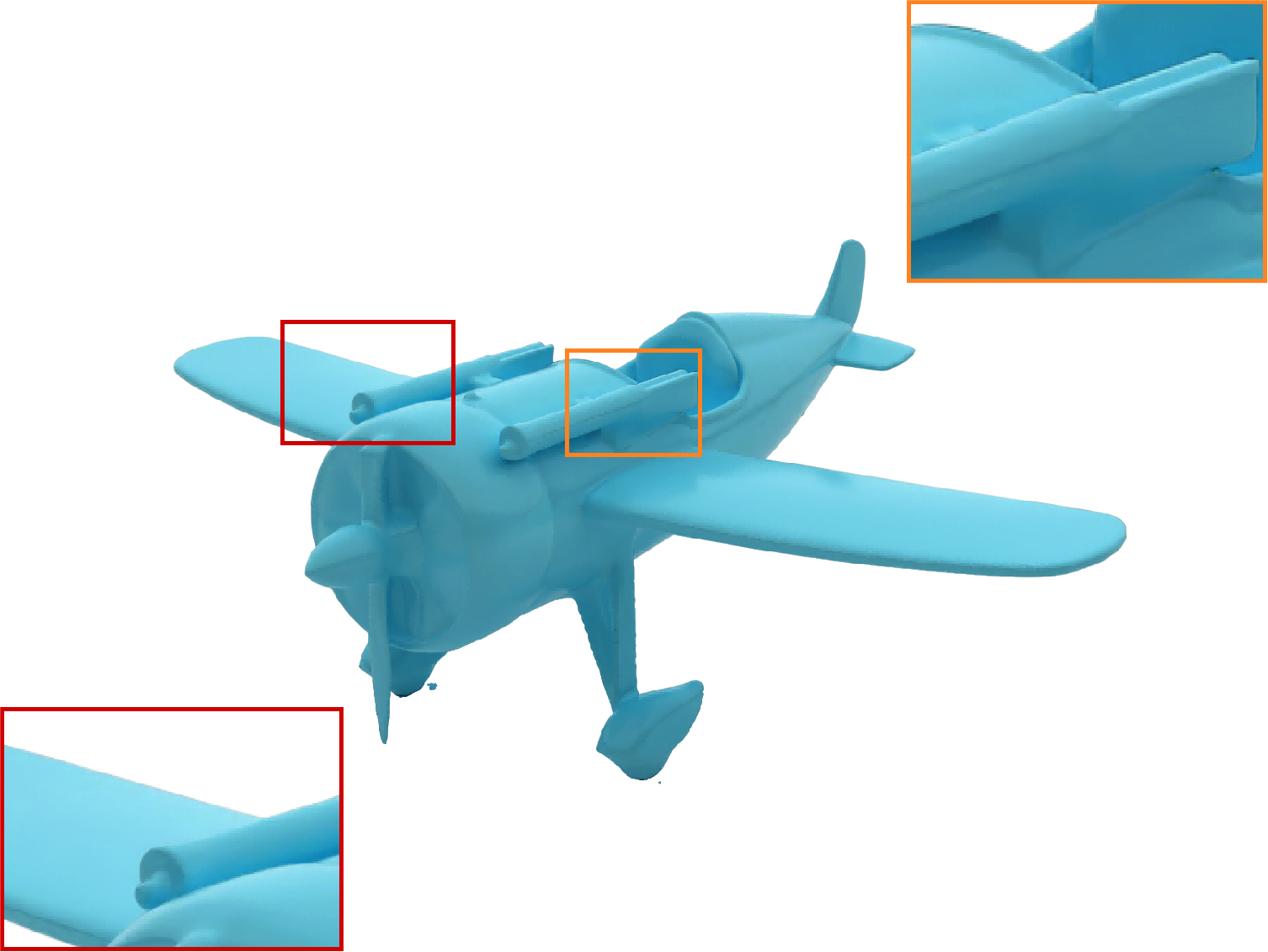}&
        \includegraphics[width=.24\textwidth]{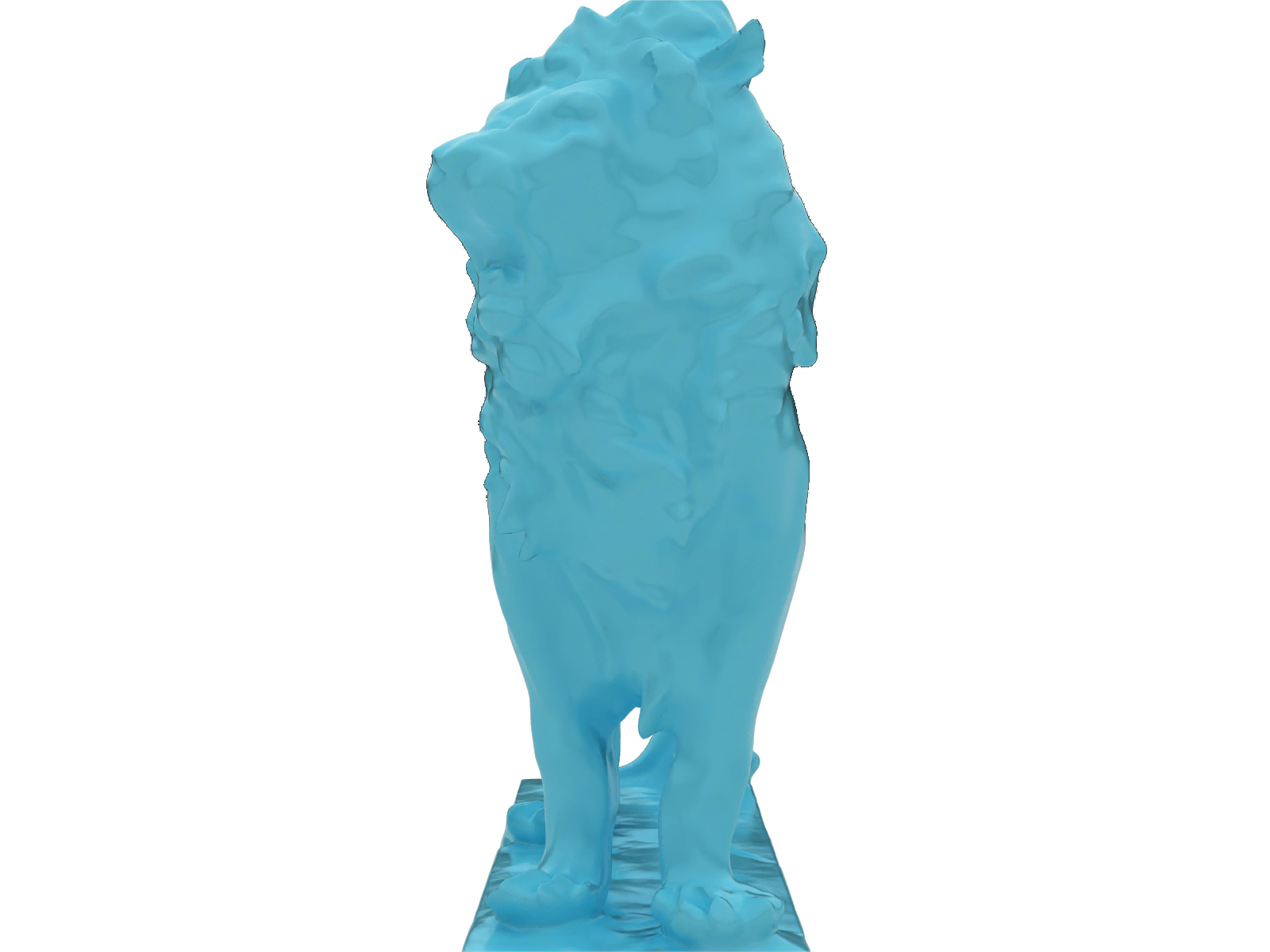} \\

    \end{tabular}
    \caption{\label{fig:neus} \textbf{Qualitative 3D Reconstruction Results.} Top: Input images; 2nd row: Ground truth; 3rd row: PGSR; Bottom: NeRO.}
\end{figure*}

\subsection{3D Shape Reconstruction}
\label{sec:3d}
\textbf{Evaluation metric.}
We adopt the same quantitative metric as NeuS~\cite{Wang21b}, calculating the Chamfer distance between point clouds to assess reconstruction accuracy. Like NeuS, we focus our evaluation on visible surface portions and exclude outliers with a Chamfer distance of over 0.15 to ensure result reliability.
We assess the performance of SOTA 3D reconstruction algorithms, including general-purpose methods (Instant-NeuS~\cite{Guo22}, NeuS2~\cite{Wang23a}, 2DGS, and PGSR), as well as NeRO~\cite{Liu23b}, designed for specular objects, and NeRRF~\cite{Chen23b}, for transparent and specular items.

\noindent
\textbf{Results and analysis.}
As shown in~\cref{tab:neus}, the algorithms perform best with Diffuse and Rough Plastic materials, achieving the lowest Chamfer distances. This shows that the uniform, view-independent reflectivity of these materials matches the algorithms' underlying assumption of diffuse surfaces, as evident in~\cref{fig:neus} with detailed reconstructions of diffuse objects. For a rigorous comparison of how different materials affect performance, see~\cref{sec:ablation_material}.

By contrast, materials such as Conductor and Dielectric, with high specular reflections and refractions, present significant challenges, resulting in substantially higher Chamfer distances. These materials, particularly in their smooth forms, are difficult to reconstruct because of their complex interactions with light, including sharp reflections and transparency. Examples of these challenges, as shown in~\cref{fig:neus}, are artifacts such as the motorbike body, containing many holes due to ground reflections causing multi-view color inconsistencies, and the camera handle, exhibiting holes resulting from transmitted light.

PGSR is the general-purpose method that performs the best in most categories. This evidences the effectiveness of the geometry regularization strategy when handling challenging reflections and transparency while relying on a diffuse physical model.
Specialized algorithms for challenging materials are far less numerous than general-purpose ones. 
Notably, NeRO outperforms the other methods on the Conductor and Rough Conductor materials, thanks to its custom-designed BRDF-based physical loss function. This function enhances its ability to simulate complex light-material interactions, especially in scenarios with view-dependent color variations caused by reflections.




\vspace{-.5em}
\subsection{Novel View Synthesis}
\textbf{Evaluation metric.} For this task, the objects are rendered in the same scene as the input but from novel viewpoints, and the renderings are compared to the ground-truth images.
We use standard quantitative metrics to evaluate the quality of the rendered 2D images: PSNR, SSIM, and LPIPS. We evaluate SOTA methods including general-purpose algorithms such as Gaussian Splatting~\cite{Kerbl23}, Instant-NGP~\cite{Muller22}, 2DGS, PGSR, and GES, as well as GSDR and GaussianShader, which are specifically designed for handling specular reflective surfaces.

\noindent\textbf{Results and analysis.}
As shown in~\cref{tab:nerf}, the results for novel view synthesis highlight the same pattern as for 3D reconstruction. Specifically, the algorithms perform best on Diffuse and Rough Plastic materials and face the greatest challenges with Conductor and Dielectric materials. Such materials
lead to chaotic color blocks appearing in the rendered images. For instance, as shown in~\cref{fig:nerf}, the metal surface of the phonograph horn reflects the chessboard pattern on the ground, making it hard for the algorithms to accurately reconstruct this part. Furthermore, the algorithms yield better results on the rough versions of these materials than on their smooth counterparts.

In terms of PSNR, the general-purpose methods Gaussian Splatting and Instant-NGP occupy the top positions in most categories on the leaderboard. This indicates that, despite their simplicity, the photometric loss functions used by these algorithms are currently the most robust to deal with various material, changes in scene lighting, and variations in object geometry. 
PGSR achieves the best results in the Conductor category, where the biggest challenge comes from specular reflections. PGSR benefits from well-designed geometrical regularization terms, which prove to be effective across a wide range of scenes and geometries despite the diffuse surface assumption. 
Specialized algorithms such as GSDR and GaussianShader do not stand out in this benchmark but will show notable performance in certain categories during subsequent ablation studies (\cref{sec:ablation_material} and \cref{sec:ablation_geo}). 
This suggests that their adaptations to the more accurate physical model are effective, yet lack robustness. In the future, combining better regularization terms with accurate physical models to enhance robustness might enable these algorithms to excel in datasets with extensive scene, material, and geometric variations, such as OpenMaterial.

\newcommand{\bigimgwidth}{.38\textwidth}
\newcommand{\smallimgwidth}{.14\textwidth}
\begin{figure*}[t]
    \setlength{\belowcaptionskip}{-5pt}
    \setlength{\tabcolsep}{1.25pt}
    \renewcommand{\arraystretch}{0.5}
    \begin{tabular}{cc}
        \hspace{.95cm}
        \begin{minipage}{\bigimgwidth}
            \includegraphics[width=\textwidth]{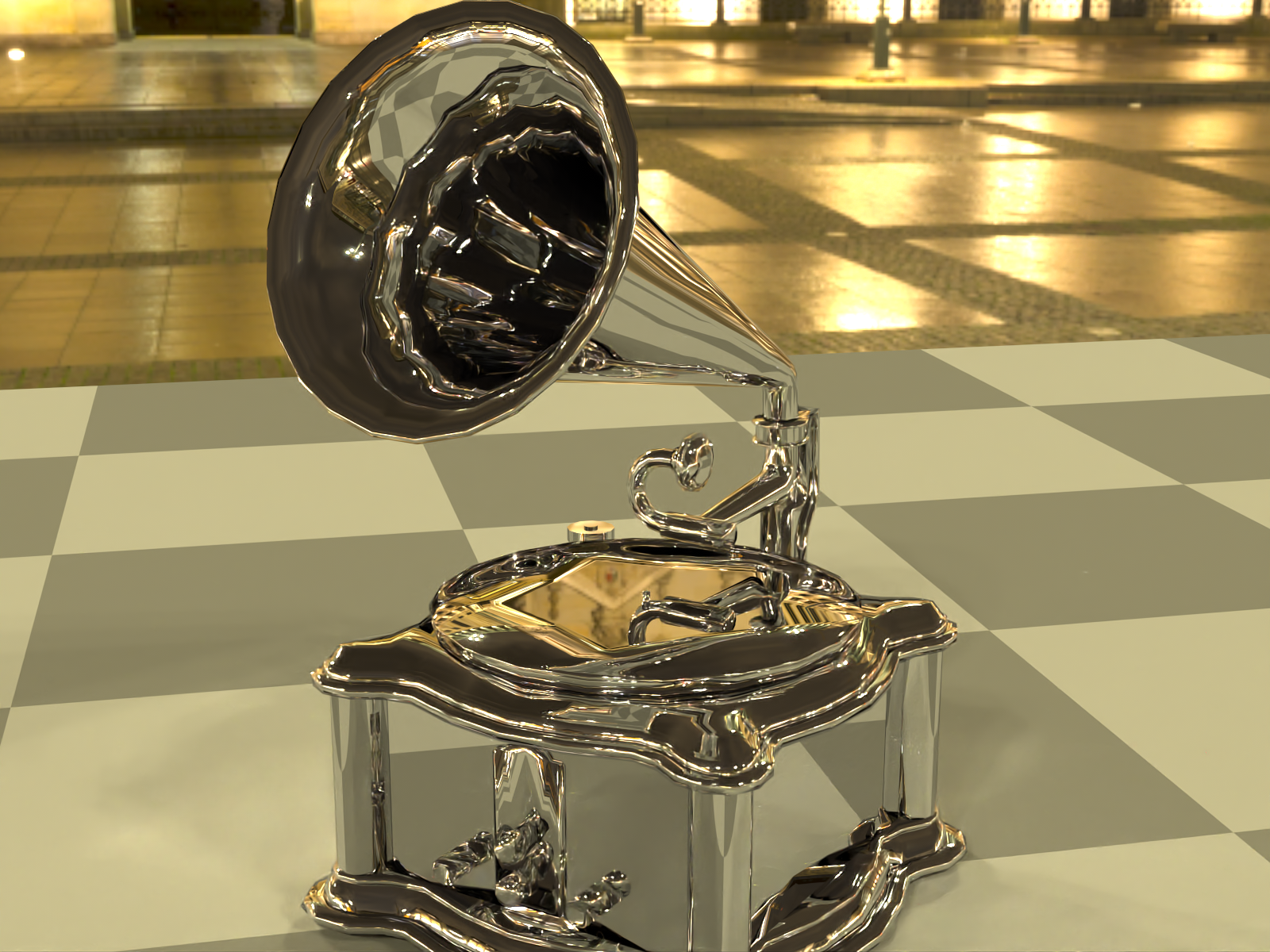}
            \raisebox{2pt}[0pt][0pt]{\makebox[\textwidth][c]{Conductor}}
        \end{minipage} &
        \hspace{7pt}
        \begin{minipage}{\smallimgwidth}
            \centering
            \begin{tabular}{ccc}
                \includegraphics[width=\textwidth]{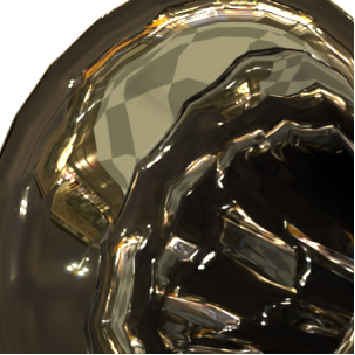} &
                \hspace{5pt}
                \includegraphics[width=\textwidth]{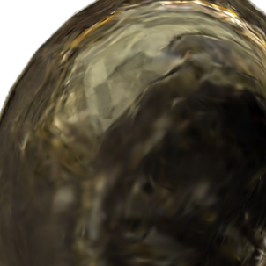} &
                \hspace{5pt}
                \includegraphics[width=\textwidth]{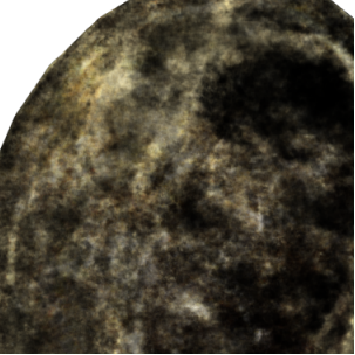} \\
                \includegraphics[width=\textwidth]{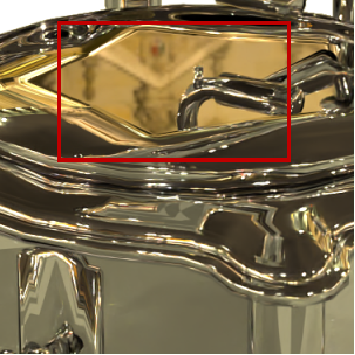} &
                \hspace{5pt}
                \includegraphics[width=\textwidth]{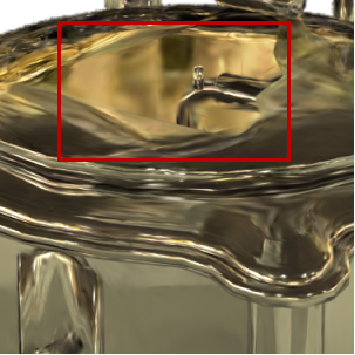} &
                \hspace{5pt}
                \includegraphics[width=\textwidth]{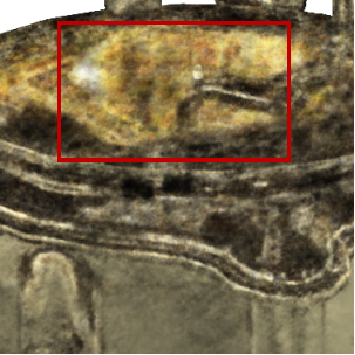} \\
                \multicolumn{1}{c}{Ground Truth} & \multicolumn{1}{c}{PGSR~\cite{Chen24}} & \multicolumn{1}{c}{InstantNGP~\cite{Muller22}}
            \end{tabular}
        \end{minipage}
    \end{tabular}
    

\caption{\label{fig:nerf} \textbf{Qualitative results for novel view synthesis.} From left to right: GT (global), GT (local), PGSR, Instant-NGP. Additional results are provided in the supplementary material.}
\end{figure*}

\begin{table*}[tb]
\centering
\scriptsize
\setlength{\tabcolsep}{3.3pt}
\setlength{\belowcaptionskip}{-12pt}
\begin{tabular}{c|c|ccccccc}
\toprule
\multirow{2}{*}{Metric} & \multirow{2}{*}{Method} & \multicolumn{7}{c}{\textbf{Material Type}} \\
                       &                         & Conductor & Dielectric & Plastic & Rough Conductor & Rough Dielectric & Rough Plastic & Diffuse \\
\midrule
\multirow{7}{*}{PSNR$\uparrow$}
                       & Gaussian Splatting~\cite{Kerbl23}    & \cellcolor{top3}22.8174 & 23.2858                 & \cellcolor{top1}36.2718 & \cellcolor{top2}30.8321 & \cellcolor{top3}30.7457 & \cellcolor{top1}39.3673 & \cellcolor{top1}41.0518 \\
                       & Instant-NGP~\cite{Muller22}           & \cellcolor{top2}22.8916 & \cellcolor{top1}24.5728 & \cellcolor{top2}36.0873 & \cellcolor{top1}31.0682 & \cellcolor{top1}32.3506 & \cellcolor{top2}38.5815                 & \cellcolor{top2}39.5141 \\
                       & 2DGS~\cite{Huang24}                             & 21.0094                 & 21.6837                 & \cellcolor{top3}34.9486 & 27.2400                 & 29.2322                 & \cellcolor{top3}37.9874                 & \cellcolor{top3}39.3214 \\
                       & PGSR~\cite{Chen24}                              & \cellcolor{top1}23.4424 & \cellcolor{top2}24.4370 & 33.6364                 & \cellcolor{top3}30.6653 & \cellcolor{top2}31.0981 & 35.5454                 & 35.7990 \\
                       & GES~\cite{Hamdi24}                              & 22.7284                 & 23.2494                 & 33.8938                 & 29.9151                 & 29.4524                 & 36.0048                 & 36.6484 \\
                       & GSDR$\star$~\cite{Ye24}                                & 22.3898                 & \cellcolor{top3}23.5641 & 34.4273                 & 29.2845                 & 29.6932                 & 36.7090                 & 37.3950 \\
                       & GaussianShader$\star$~\cite{Jiang23}                   & 18.6242                 & 21.2597                 & 23.1657                 & 20.7524                 & 23.4626                 & 23.2146                 & 23.2493 \\
\midrule
\multirow{7}{*}{SSIM$\uparrow$}
                       & Gaussian Splatting    & 0.8500                  & 0.8255                  & 0.9703                  & 0.9459                 & 0.9408                 & 0.9819                 & 0.9865 \\
                       & Instant-NGP           & 0.8481                  & \cellcolor{top2}0.8473  & \cellcolor{top3}0.9763  & \cellcolor{top3}0.9526 & \cellcolor{top2}0.9591 & \cellcolor{top3}0.9880 & \cellcolor{top1}0.9911 \\
                       & 2DGS                  & 0.8099                  & 0.7973                  & 0.9706                  & 0.9155                 & 0.9294                 & 0.9871                 & \cellcolor{top2}0.9905 \\
                       & PGSR                  & \cellcolor{top1}0.8770  & \cellcolor{top1}0.8585  & \cellcolor{top1}0.9785  & \cellcolor{top1}0.9639 & \cellcolor{top1}0.9686 & \cellcolor{top2}0.9890 & 0.9899 \\
                       & GES                   & \cellcolor{top3}0.8528  & 0.8300                  & 0.9692                  & 0.9468                 & 0.9422                 & 0.9807                 & 0.9848 \\
                       & GSDR $\star$                 & \cellcolor{top2}0.8697  & \cellcolor{top3}0.8449  & \cellcolor{top2}0.9782  & \cellcolor{top2}0.9584 & \cellcolor{top3}0.9574 & \cellcolor{top1}0.9894 & \cellcolor{top2}0.9905 \\
                       & GaussianShader$\star$        & 0.8118                  & 0.8197                  & 0.9133                  & 0.8738                 & 0.9139                 & 0.9246                 & 0.9253 \\
\midrule
\multirow{7}{*}{LPIPS$\downarrow$}
                       & Gaussian Splatting    & \cellcolor{top2}0.1152 & \cellcolor{top2}0.1261 & \cellcolor{top2}0.0386 & \cellcolor{top1}0.0530 & \cellcolor{top1}0.0696 & \cellcolor{top2}0.0204 & \cellcolor{top1}0.0152 \\
                       & Instant-NGP           & 0.1523                 & 0.1464                 & 0.0591                 & 0.0848                 & 0.0910                 & 0.0382                 & \cellcolor{top3}0.0242 \\
                       & 2DGS                  & 0.1313                 & \cellcolor{top3}0.1348 & \cellcolor{top1}0.0378 & 0.0720                 & \cellcolor{top3}0.0718 & \cellcolor{top1}0.0201 & \cellcolor{top2}0.0155 \\
                       & PGSR                  & \cellcolor{top1}0.1112 & \cellcolor{top1}0.1192 & \cellcolor{top3}0.0519 & \cellcolor{top2}0.0594 & \cellcolor{top2}0.0708 & 0.0350                 & 0.0318 \\
                       & GES                   & \cellcolor{top3}0.1232 & \cellcolor{top3}0.1348 & 0.0538                 & \cellcolor{top3}0.0644 & 0.0817                 & 0.0361                 & 0.0319 \\
                       & GSDR$\star$                  & 0.1239                 & 0.1400                 & 0.0522                 & 0.0687                 & 0.0911                 & \cellcolor{top3}0.0337 & 0.0315 \\
                       & GaussianShader$\star$        & 0.1863                 & 0.1841                 & 0.1525                 & 0.1654                 & 0.1646                 & 0.1434                 & 0.1424 \\
\bottomrule
\end{tabular}
\caption{\textbf{Comparison of SOTA Novel View Synthesis Methods.} A $\star$ denotes specialized methods.}
\label{tab:nerf}
\end{table*}

\subsection{Ablation Study}

In addition to our full OpenMaterial dataset, we have designed a controlled dataset to analyze the factors impacting 3D reconstruction and novel view synthesis. Specifically, we selected five objects with varying complexities (Vase, Snail, Boat, Motor Bike, and Statue), and three distinct lighting conditions (an indoor scene, a daytime garden, and a nighttime street). Combined with our seven material types, this yields 105 unique scenes (5 objects × 3 lighting conditions × 7 materials). This setup is designed to evaluate the performance of SOTA algorithms under controlled variables: Material properties, shape geometry, and lighting conditions. Below, we provide the key results of our analysis; lighting conditions part, additional visualizations, and detailed metrics are given in the supplementary materials.

\subsubsection{Material Properties}
\label{sec:ablation_material}
To highlight the impact of material properties on 3D reconstruction and novel view synthesis, we controlled shape and lighting conditions, focusing our ablation study on material properties. The results in~\cref {tab:ablation_material} confirm the previous findings in~\cref {sec:3d}: Existing algorithms excel with diffuse materials such as Diffuse and Rough Plastic and struggle with specular materials such as Conductor and Dielectric.

Nevertheless, the ranking of the algorithms for novel view synthesis differ from that on the complete OpenMaterial dataset (\cref{tab:nerf}), which encompasses diverse geometry shapes, lighting, and material types.
Specifically, the general-purpose methods 2DGS, PGSR, and GES are higher in the ranking, and so is the specifically-designed GSDR algorithm. This suggests that the efforts of 2DGS and GES in finding better feature representations are effective, and GSDR's efforts to better integrate physical models are also paying off, although robustness remains limited. 
PGSR's consistently good ranking on both leaderboards reaffirms the effectiveness of its carefully designed regularization term. 

\begin{table*}[tb]  
\centering  
\scriptsize  

\setlength{\belowcaptionskip}{-15pt}
\setlength{\tabcolsep}{3pt}  
\begin{tabular}{p{1.cm}|c|ccccccc}  
\toprule  
\multirow{2}{*}{\parbox{1.cm}{\centering Metric}} & \multirow{2}{*}{Method} & \multicolumn{7}{c}{\textbf{Material Type}} \\  
                       &                         & Conductor & Dielectric & Plastic & Rough Conductor & Rough Dielectric & Rough Plastic & Diffuse \\  
\midrule  
\multirow{5}{*}{\parbox{1.cm}{\centering Chamfer Distance$\downarrow$}}
                       & 2DGS       & \num{2.0784}                 & \num{1.8583}                 & \cellcolor{top3}\num{0.6116}     & \num{1.7614}                    & \num{1.6913}                        & \cellcolor{top3}\num{0.5903}    & \cellcolor{top3}\num{0.5565} \\
                       & Instant-NeuS              & \num{1.469488}                 & \cellcolor{top3}\num{1.335089} & \num{0.652197}                     & \num{1.392668}                    & \cellcolor{top3}\num{1.287987}        & \num{0.635543}                    & \num{0.610307} \\
                       & NeuS2                     & \cellcolor{top3}\num{1.4161} & \cellcolor{top2}\num{1.1342} & \num{0.6483}                     & \cellcolor{top3}\num{1.2518}    & \cellcolor{top2}\num{1.0931}        & \num{0.6322}                    & \num{0.6256} \\
                       & PGSR                      & \cellcolor{top2}\num{1.3137} & \cellcolor{top1}\num{0.9845} & \cellcolor{top1}\num{0.5063}     & \cellcolor{top2}\num{1.0403}    & \cellcolor{top1}\num{0.9267}        & \cellcolor{top1}\num{0.4807}    & \cellcolor{top2}\num{0.4814} \\
                       & NeRO$\star$                      & \cellcolor{top1}\num{0.7683} & \num{1.9700}                 & \cellcolor{top2}\num{0.5523}     & \cellcolor{top1}\num{0.7483}    & \num{1.5726}                        & \cellcolor{top2}\num{0.5514}    & \cellcolor{top1}\num{0.4710} \\
\midrule
\multirow{7}{*}{\parbox{1.cm}{\centering PSNR$\uparrow$}}        
                       & Gaussian Splatting      & \cellcolor{top2}\num{21.3856}    & \num{21.926318} & \cellcolor{top3}\num{34.161479}     & \num{28.112349}      & \num{28.425814}  & \cellcolor{top3}\num{35.124162}  & \cellcolor{top3}\num{35.269029} \\
                       & Instant-NGP  & \num{21.038271} & \cellcolor{top2}\num{23.396643} & \num{34.104555} & \num{27.093283} & \cellcolor{top1}\num{30.985292} & \num{34.714689} & \num{34.771471}\\
                       & 2DGS                    & \num{19.926959}  &  \num{20.41677} &  \cellcolor{top1}\num{35.595247}    &  \num{24.851087}     &  \num{28.475166}  &  \cellcolor{top1}\num{36.523827}  & \cellcolor{top2}\num{37.182538}\\
                       & PGSR                    & \cellcolor{top3}\num{21.286408}  & \cellcolor{top1}\num{23.93008}  & \num{33.754025}     & \cellcolor{top1}\num{29.423777}      & \cellcolor{top2}\num{30.400894}  & \num{34.397354}  & \num{34.456235} \\
                       & GES                     & \cellcolor{top1}\num{21.5051}   & \num{22.2645}  & \num{34.1418}   & \cellcolor{top2}\num{28.4124}      & \num{28.6211}    & \num{35.0918}  & \num{35.1559} \\
                       & GSDR$\star$         & \num{21.278171}  & \cellcolor{top3}\num{22.729077} & \cellcolor{top2}\num{34.956552}     & \cellcolor{top3}\num{28.135336}      & \cellcolor{top3}\num{29.295549}  &  \cellcolor{top2}\num{36.148513}  & \cellcolor{top1}\num{37.36471} \\ 
                       & GaussianShader$\star$           & \num{18.065133} & \num{20.508144} & \num{23.760391} & \num{20.608701} & \num{23.284909} & \num{23.368272} & \num{23.994903} \\
\bottomrule  
\end{tabular}  
\caption{\textbf{Ablation of Material Properties.}}  
\label{tab:ablation_material}
\end{table*}

\begin{wraptable}{r}{0.6\linewidth}
\vspace{-10pt}
\centering
\setlength{\belowcaptionskip}{-10pt}
\scriptsize  
\setlength{\tabcolsep}{2.pt}  

\begin{tabular}{p{.7cm}|c|ccccc}  
\toprule  
\multirow{2}{*}{\parbox{.65cm}{\centering Metric}} & \multirow{2}{*}{Method} & \multicolumn{5}{c}{\textbf{Object Name}} \\  
                       &                         & Vase & Snail & Boat & Motor Bike & Statue  \\  
\midrule  
\multirow{5}{*}{\parbox{.65cm}{\centering CD$\downarrow$}}  
                       & 2DGS       & \num{1.8806}                  & \num{2.1451} & \cellcolor{top1}\num{0.7352}   & \num{1.0741}                   & \num{1.0600}                  \\
                       & Instant-NeuS              & \cellcolor{top2}\num{1.236101}  & \num{1.449687} & \cellcolor{top3}\num{1.006113}   & \cellcolor{top3}\num{0.965645}   & \num{0.8034}                  \\
                       & NeuS2                     & \num{1.3190}                  & \cellcolor{top1}\num{1.0958} & \num{1.0867}                   & \cellcolor{top2}\num{0.9655}   & \cellcolor{top3}\num{0.5492}  \\
                       & PGSR                     & \cellcolor{top3}\num{1.2407}  & \cellcolor{top3}\num{1.1155}  & \cellcolor{top2}\num{0.8168}   & \cellcolor{top1}\num{0.8779}   & \cellcolor{top1}\num{0.2995}  \\
                       & NeRO$\star$                      & \cellcolor{top1}\num{1.2317}  & \cellcolor{top2}\num{1.0984} & \num{1.1276}                   & \num{1.0740}                   & \cellcolor{top2}\num{0.5159}  \\
\midrule
\multirow{7}{*}{\parbox{.65cm}{\centering PSNR$\uparrow$}}  
                       & Gaussian Splatting          & \num{28.855678}  & \num{29.161615} & \num{29.409573}  & \cellcolor{top3}\num{28.781895}  & \num{29.794633} \\
                       & Instant-NGP  & \cellcolor{top1}\num{29.962749}  & \cellcolor{top3}\num{30.145689}  & \num{28.940783}  & \num{28.38213}  & \num{29.785938} \\
                       & 2DGS                    & \num{24.921071} & \cellcolor{top1}\num{30.796852}  & \cellcolor{top2}\num{30.436085} & \num{28.564278} & \cellcolor{top3}\num{30.261425} \\
                       & PGSR                    & \cellcolor{top2}\num{29.244217} & \cellcolor{top2}\num{30.232075} & \num{29.131888} & \cellcolor{top2}\num{29.223300} & \cellcolor{top2}\num{30.489071} \\
                       & GES & \cellcolor{top3}\num{29.047588} & \num{29.289801} & \cellcolor{top3}\num{29.700217} & \num{28.70121} & \num{29.827252} \\
                       & GSDR$\star$        & \num{28.795018} & \num{29.994795}  & \cellcolor{top1}\num{30.693927} & \cellcolor{top1}\num{29.373514} & \cellcolor{top1}\num{31.076965} \\  
                       & GaussianShader$\star$ & \num{22.848308} & \num{21.208088} & \num{22.678301} & \num{20.344479} & \num{22.62829} \\
\bottomrule  
\end{tabular}  

\caption{\textbf{Ablation of Geometric Shape.}}  
\label{tab:ablation_shape}

\end{wraptable}

\subsubsection{Geometric Shape}
\label{sec:ablation_geo}
To highlight the impact of geometric shape on 3D reconstruction and novel view synthesis, we controlled the material and lighting conditions, focusing our ablation study exclusively on shape.
Overall, the results in~\cref{tab:ablation_shape} suggest that 3D reconstruction is more strongly influenced by 3D shape than novel view synthesis. In 3D reconstruction, the algorithms generally perform better on statues than on vases. We attribute this to the more detailed surface structures of statues, which aid the algorithms in pinpointing 3D positions. This also underscores the challenges that algorithms face when dealing with large, low-texture, and smooth areas. On vases, algorithms such as NeRO, Instant-NeuS, and PGSR perform well, indicating that enhancing the underlying physical model and utilizing regularization terms to leverage prior information can improve results on low-texture and smooth areas. In novel view synthesis, the specially-designed algorithm GSDR yields the best results across multiple objects, validating the effectiveness of modifying the underlying physical model. However, as mentioned in~\cref{sec:ablation_material}, GSDR lacks robustness; thus, improvements in this area, such as designing better regularization terms, could potentially improve its performance.

\section{Limitations and Conclusion}
\label{sec:limitations_and_conclusion}
We have introduced OpenMaterial, the first large-scale dataset designed for quantitative assessment of 3D reconstruction methods for objects with complex materials, featuring an extensive variety of shapes, material types, and lighting conditions. 
Our dataset not only supports extensive evaluations but also facilitates the analysis of the capabilities of existing algorithms at handling challenging materials. We have validated its utility by evaluating SOTA 3D reconstruction and novel view synthesis algorithms. 
The main limitation of our approach, however, lies in the dataset expansion. While it is possible to increase the variety of shapes by incorporating models from existing open-source datasets, the quality of many models does not meet real-world standards. This requires manual selection to ensure realism, a process that could become a bottleneck if demands for larger data volumes arise.
\clearpage
\bibliographystyle{IEEEtran}
\bibliography{bibtex/string,bibtex/vision,bibtex/physics,bibtex/graphic}


\end{document}